\documentclass{article} 
\usepackage[dvipsnames]{xcolor}
\usepackage{subcaption}
\usepackage{pgf}
\usepackage{adjustbox}
\usepackage{wrapfig}
\usepackage{tikz}
\usepackage{enumitem}
\usepackage{tcolorbox}
\usepackage{comment}
\usepackage{titletoc}
\usepackage{booktabs}
\usepackage{algorithm}
\usepackage{algpseudocode}
\usepackage{amsmath}
\usepackage{amssymb}
\usepackage{natbib}
\usepackage[table]{xcolor}
\usepackage[colorlinks=true, citecolor=black]{hyperref}
\usepackage{booktabs}
\usepackage{adjustbox}
\usepackage[table]{xcolor}
\hypersetup{colorlinks=true,allcolors=[rgb]{0.0,0.1,0.5}}
\NewDocumentCommand{\incplt}{O{\columnwidth}m}{%
  \begin{center}
    \adjustbox{center}{\adjustbox{width=#1+10pt}{\includegraphics[width=#1]{./figures/#2.pdf}}}
  \end{center}
}
\usepackage{iclr2027_conference,times}

\usepackage{amsmath,amsfonts,bm}

\def\eqref#1{equation~\ref{#1}}

\def\1{\bm{1}}

\DeclareMathAlphabet{\mathsfit}{\encodingdefault}{\sfdefault}{m}{sl}
\SetMathAlphabet{\mathsfit}{bold}{\encodingdefault}{\sfdefault}{bx}{n}

\DeclareMathOperator*{\argmax}{arg\,max}

\usepackage{url}

\title{Deep Epistemic Value Functions \\ for Optimistic Exploration}

\author{%
  Leander Diaz‐Bone\thanks{Correspondence to Leander Diaz‐Bone (ldiazbone@ethz.ch).}\textsuperscript{\normalfont \;\;,1,2}%
  \quad Marco Bagatella\textsuperscript{\normalfont 1,2}%
  \quad Jonas Hübotter\textsuperscript{\normalfont 1}%
  \quad Andreas Krause\textsuperscript{\normalfont 1}%
  \\[3pt]
  \textsuperscript{1}ETH Zürich, Switzerland
  \quad \textsuperscript{2}Max Planck Institute for Intelligent Systems, Germany
  \ifarxiv
  \\[1.5ex]
  \href{https://github.com/LeanderDiazBone/devote}{    \texttt{https://github.com/LeanderDiazBone/devote}
  }
  \fi
}

\newif\ifarxiv

\arxivtrue 

\ifarxiv
  \iclrfinalcopy
\fi
\begin{document}

\maketitle
\ifarxiv
  \fancyhead{}
  \fancyhead[L]{\footnotesize Preprint. Under review.}
\fi

\begin{abstract}
Principled exploration in reinforcement learning requires an agent to quantify its epistemic uncertainty and act to resolve it.
Uncertainty over the value function provides a natural signal for exploration, yet existing deep approximations remain brittle and perform inconsistently.
The central challenge is therefore to scale these ideas robustly.
We conduct a systematic empirical study of how epistemic uncertainty is represented, propagated, and optimized in deep epistemic value functions, and uncover distinct failure modes along each of these axes.
These findings motivate DEVOTE, a model-free reinforcement learning algorithm that controls how uncertainty generalizes beyond observed data, stabilizes its temporal propagation, and preserves adaptation to the resulting non-stationary exploration objective.
Across reward-free exploration and challenging continuous-control tasks, DEVOTE reaches novel states more effectively and achieves higher task return than strong model-free and model-based exploration baselines.
These results provide evidence that deep epistemic value functions are a promising path toward scalable, principled exploration.
\end{abstract}

\begin{figure}[h]
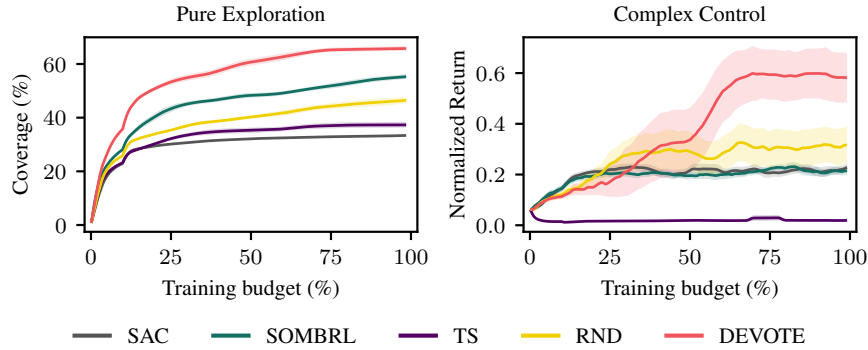

    \centering
    \vspace{-3ex}
    \incplt[0.8\textwidth]{figures_8pt/summary}
    \vspace{-3ex}
    \caption{DEVOTE provides stable and scalable estimates of a deep epistemic value function, guides the exploratory policy toward novel and informative states, and outperforms strong RL exploration methods across a range of environments. We plot the average coverage and performance across environments outlined in Section~\ref{sec:results}.}
    \label{fig:}
    \vspace{-3ex}
\end{figure}

\section{Introduction}

Deep reinforcement learning (RL) enables agents to acquire complex behaviors through interaction, yet current methods remain notoriously sample inefficient, often requiring millions of environment interactions \citep{dulac-arnold_challenges_2019}. 
A central reason is exploration: practical deep RL algorithms still largely rely on local, temporally independent action perturbations, which cannot sustain coherent exploration over long horizons and may require time exponential in the horizon to discover rewarding behavior \citep{osband2016generalizationexplorationrandomizedvalue}. 
Sample-efficient RL, in contrast, requires purposeful exploration: the agent should seek out informative parts of the state space and reduce uncertainty about how to act optimally.
Theoretical work --- often developed in the context of bandits --- formalizes this through uncertainty-directed strategies such as optimism in the face of uncertainty \citep{Srinivas_2012} and Thompson sampling \citep{russo2020tutorialthompsonsampling}. 
Translating these principles to deep RL, however, remains challenging due to the difficulty of reliable uncertainty quantification with deep neural networks \citep{abdar_review_2021} and the sensitivity of deep RL algorithms to implementation choices \citep{agarwal_deep_2022}.

We argue that deep epistemic value functions --- which quantify epistemic uncertainty about the long-term consequences of the policy's actions --- are a promising path toward scalable, sample-efficient RL, because they represent the long-term objective that policies optimize. 
Value-directed exploration also admits strong theoretical guarantees \citep{osband2016generalizationexplorationrandomizedvalue, ishfaq_randomized_2021}. 
The central challenge is to realize these principles robustly under deep function approximation.
Scaling deep epistemic value functions therefore requires understanding how to \textbf{represent}, \textbf{propagate}, and \textbf{optimize} epistemic uncertainty in practice.

In this work, we conduct a systematic empirical study of deep epistemic value functions and uncover distinct challenges along each of these three dimensions.
We find that uncertainty can collapse beyond observed data, temporal-difference learning can destabilize its propagation over time, and the continually changing exploration objective can cause the critic and policy to lose the plasticity required to keep adapting. 
We address these failures by explicitly controlling how uncertainty generalizes beyond the data, separating epistemic propagation from reward learning, and preserving the plasticity needed to track a continually moving exploration objective.
Combining these ideas, we introduce \textit{D}eep \textit{E}pistemic \textit{V}alue functions for \textit{O}p\textit{T}imistic \textit{E}xploration (DEVOTE), which improves exploration consistently across a diverse set of environments.

Our contributions are:
\begin{enumerate}[leftmargin=10pt, itemsep=1pt, topsep=0pt]
    \item We conduct a systematic empirical study of deep epistemic value functions and identify practical challenges in reliably representing, propagating, and optimizing epistemic uncertainty with deep function approximation.
    \item Building on these insights, we develop targeted remedies and combine them in DEVOTE, a model-free algorithm for scalable optimistic exploration with deep epistemic value functions.
    \item We show that DEVOTE outperforms strong model-free and model-based exploration baselines across challenging benchmarks, and use ablations to isolate the contribution of each component.
\end{enumerate}

\section{The Exploration Problem}

We consider the infinite-horizon discounted reinforcement learning setting \citep{sutton2018reinforcement}, formalized as a Markov decision process (MDP) $\mathcal{M} = (\mathcal{S}, \mathcal{A}, p, \mu_0, r, \gamma)$. Here, $\mathcal{S}$ and $\mathcal{A}$ are the state and action spaces, $p \colon \mathcal{S} \times \mathcal{A} \to \Delta(\mathcal{S})$ is the transition kernel, $\mu_0 \in \Delta(\mathcal{S})$ is the initial state distribution, $r \colon \mathcal{S} \times \mathcal{A} \to \mathbb{R}$ is the reward function, and $\gamma \in (0,1)$ is the discount factor.
A stationary policy $\pi \colon \mathcal{S} \to \Delta(\mathcal{A})$ and $\mathcal{M}$ together induce a distribution $\rho^\pi$ over trajectories, generated by $s_0 \sim \mu_0$, $a_t \sim \pi(\cdot \mid s_t)$, $r_t = r(s_t, a_t)$, and $s_{t+1} \sim p(\cdot \mid s_t, a_t)$.
The $Q$-function of $\pi$ measures the expected discounted return after taking action $a$ in state $s$,
\begin{equation}
    Q^\pi(s,a) = \mathbb{E}_{\rho^\pi}\Big[\sum_{t=0}^{\infty} \gamma^t r_t \,\Big|\, s_0 = s, a_0 = a\Big],
    \label{eq:q-function}
\end{equation}
and is the unique fixed point of the Bellman equation $Q^\pi(s,a) = \mathbb{E}_{p, \pi}\big[r(s,a) + \gamma Q^\pi(s',a')\big].$
The value of $\pi$ in state $s$ is $V^\pi(s) = \mathbb{E}_{a \sim \pi(\cdot \mid s)}[Q^\pi(s,a)]$. 
The agent seeks a policy that maximizes the expected return $J(\pi) = \mathbb{E}_{s \sim \mu_0}[V^\pi(s)]$, and we write $\pi^\star \in \argmax_\pi J(\pi)$ for an optimal policy.\par
We consider the episodic setting, where in episode $n$ the agent rolls out policy $\pi_n$ in the unknown MDP $\mathcal{M}$, observes transitions and rewards, and updates its behavior from the collected data.
Because this data is generated by the agent's own actions, sample-efficient learning requires purposeful exploration to collect information that can resolve uncertainty about how to act optimally.
The optimal Q-function $Q^\star = Q^{\pi^\star}$ determines the optimal action through $\pi^\star(s) \in \arg\max_{a \in \mathcal{A}} Q^\star(s,a)$, making uncertainty about $Q^\star$ directly relevant to decision making.
We therefore focus on deep epistemic value functions --- deep neural networks that estimate a value function together with its epistemic uncertainty --- as a promising approach to scalable and efficient exploration.
We provide further background, particularly on how $Q^\star$ may be estimated, in Appendix~\ref{app:ext_back_mf_rl}.


\section{Addressing Limitations of Deep Epistemic Value Functions} \label{sec:limitations}
Despite their promise, implementations of deep epistemic value functions remain brittle across environments.
Existing approaches typically use ensembles, optionally with randomized prior functions, to estimate epistemic uncertainty and derive exploration signals \citep{osband2016deepexplorationbootstrappeddqn, osband2018randomizedpriorfunctionsdeep, chen_ucb_2017, ciosek_better_2019, nikolov_information-directed_2019}.
We demonstrate that this gap arises from difficulties in robustly (i) estimating, (ii) propagating, and (iii) optimizing epistemic value uncertainty with deep neural networks.
In the following, we study failures along each of these axes and introduce targeted remedies, starting with general uncertainty quantification before turning to learning dynamics specific to temporal-difference value learning.

\subsection{Estimation: Calibrating Epistemic Neural Networks}
We represent deep epistemic value functions using Epistemic Neural Networks (ENNs) \citep{osband2023epistemicneuralnetworks}, a general framework for epistemic uncertainty quantification with neural networks.
An ENN $e_\theta = f_\theta + g$ combines a trainable network
$f_\theta$ with a fixed network $g$ additively in function space, and conditions its prediction on an epistemic index $z \sim p_z$, inducing a distribution over plausible outputs that represents the model's epistemic uncertainty.
Training amounts to minimizing the mean squared error in expectation over indices,
\begin{align}
    \mathcal{L}^\text{ENN}(\theta) = \mathbb{E}_{(x,y)\sim\mathcal{D}, z\sim p_z}[((f_\theta+g)(x,z) - y)^2].
\end{align}

\begin{figure}[t]
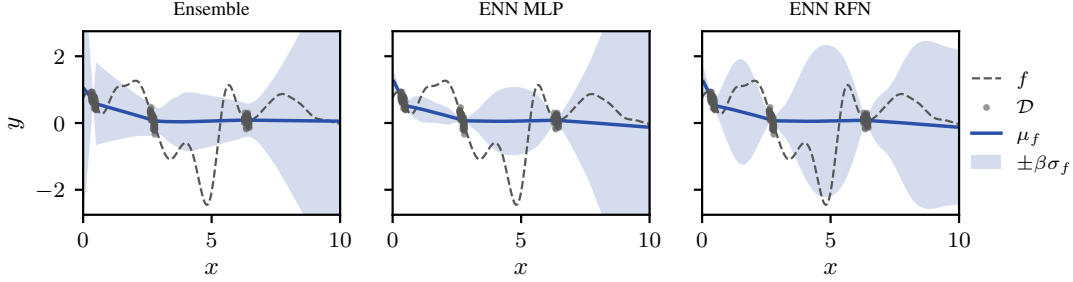

    \centering
    \vspace{-3ex}
    \incplt[\textwidth]{calibration_enns_new/enn_uncertainty_synth_gp_1_rbf_0}
    \vspace{-3ex}
    \caption{Comparison of uncertainty estimates from an Ensemble, an ENN with MLP prior (ENN MLP), and an ENN with RFN prior (ENN RFN) on a synthetic regression task. Dashed curves show the true function, dots indicate training observations, solid curves show predictive means, and shaded bands represent $\pm \beta \sigma_f$. ENNs with RFN function-space priors yield better calibrated uncertainty.}
    \label{fig:gp-calibration}
    \vspace{-3ex}
\end{figure}

The prior $g$ fixes an initial spread over plausible functions.
During training, $f_\theta(x,z)$ absorbs the residual $y - g(x,z)$ on the data for every index, collapsing the uncertainty.
Away from the data, the retained spread depends on how much the observations
determine $g$ elsewhere.

\textbf{Randomized Fourier Networks\;}
The choice of the prior $g$ is therefore critical for calibration and sample efficiency, yet in practice randomly initialized ReLU networks have been used predominantly \citep{osband2018randomizedpriorfunctionsdeep,osband2023epistemicneuralnetworks}. 
This conceptually corresponds to a prior in weight space, and the distribution over functions it induces is not well understood: we will show that it yields poorly calibrated uncertainty in the regression setting. 
We therefore propose specifying the prior in function space, where a kernel states directly how the prior $g$ reacts to changes in the input, and hence how uncertainty collapses away from the data.
When the $l_2$ distance provides a meaningful notion of similarity, we use an RBF kernel $k(x,x') = \exp(-\lVert x - x'\rVert^2 / 2\ell^2)$, whose lengthscale $\ell$ controls the variability of the prior, and can be tuned or set according to domain knowledge. 
To realize such priors in an ENN we propose Randomized Fourier Networks (RFNs), single-layer random networks that approximate a Gaussian process with a stationary kernel $k$ \citep{rahimi2007randomfourierfeatures},
\begin{align}
    g_{\text{rfn}}(x,z) = \sqrt{\frac{2}{m}}\, \mathbf{1}^\top \cos(\Omega_z x + b_z),
\end{align}
where the cosine is applied elementwise and, for each $i \in [m]$, the bias
$b_{z,i} \sim \mathcal{U}([0,2\pi])$ and the frequency row $\Omega_{z,i}$ are drawn i.i.d.\ from the spectral density of $k$.

\textbf{Function-space priors improve calibration\;}
\looseness -1 We empirically investigate how the choice of prior qualitatively determines the uncertainty an ENN produces.
We train an ensemble with $10$ members, an ENN with a standard MLP prior ($g_{\text{MLP}}(x,z)=f_{\theta_z}(x)$, where $f_{\theta_z}$ is a two-layer ReLU MLP and $p_z=\mathcal{U}([10])$ \citep{osband2023epistemicneuralnetworks}) and an ENN with an RFN prior ($g_{\text{rfn}}, p_z=\mathcal{U}([10])$) on a synthetic GP regression task in Figure~\ref{fig:gp-calibration} and the standard UCI datasets in Table~\ref{tab:calibration_main}.
On the synthetic regression task both the ensemble and the ENN with MLP prior collapse the uncertainty between the training clusters, while the RFN prior retains uncertainty that grows with distance from the data. 
The MLP prior, in contrast, is approximately linear across the domain and does not spread its variability in this way (see Figure~\ref{fig:prior-comparison}). 
We discuss the implementation details along with further results in Appendices~\ref{app:enn_cal_iml} and ~\ref{app:regression} respectively.

\subsection{Propagation: Stabilizing Epistemic Value Estimation}
A well-behaved local uncertainty estimate is insufficient for exploration unless it can be propagated reliably over time. 
We therefore now turn to using ENNs to parameterize a value function $Q_\theta(s,a,z)=(f_\theta+g)(s,a,z)$.
Value estimation is fundamentally harder than regression, as targets depend on the estimate itself (see Appendix~\ref{app:ext_back_mf_rl}).
Bootstrapping combined with off-policy data and function approximation is already known to destabilize value estimation \citep{sutton2018reinforcement}, and we find that ENNs trained naively by temporal difference (TD) learning aggravate this problem.
Writing out the TD loss under the ENN parametrization exposes this issue:
\begin{align}
    \mathcal{L}^\text{TD}(\theta) &= \mathbb{E}_{(s,a,r,s')\sim\mathcal{B},z\sim p_z,a'\sim\pi(s')}\big[\big(r + \gamma (f_{\bar{\theta}} + g)(s',a',z) - (f_\theta + g)(s,a,z)\big)^2\big] \label{eq:td-enn}\\
    &= \mathbb{E}_{\mathcal{B},p_z,\pi}\big[\big(\tilde{r} + \gamma f_{\bar{\theta}}(s',a',z) - f_\theta(s,a,z)\big)^2\big], \label{eq:td-augmented}
\end{align}
where barred parameters denote Polyak-averaged target networks, $\mathcal{B}$ is the replay buffer, and $\tilde{r} = r + \gamma g(s',a',z) - g(s,a,z)$.
The trainable network is therefore fit by standard TD-learning on an augmented reward that mixes the reward with the prior difference.
As $g$ is a fixed random function, uncorrelated with the reward, this difference injects high-variance noise into the Bellman target, which bootstrapping propagates through the value estimate: the prior leaks into the reward signal and destabilizes value estimation.

\begin{figure}[t]
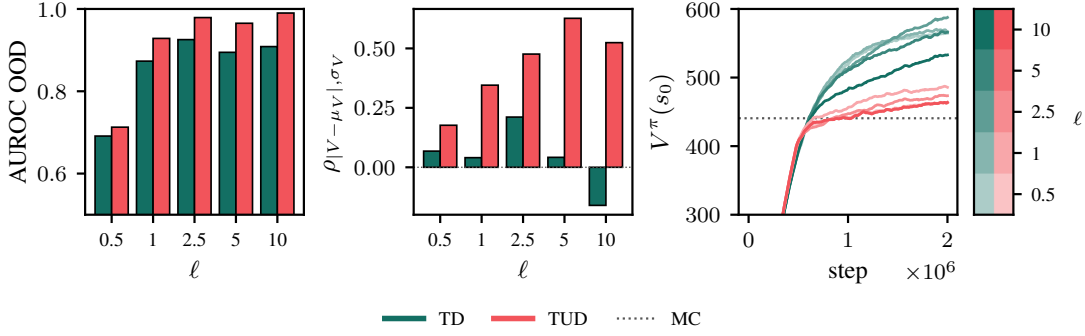

    \centering
    \vspace{-3ex}
    \incplt[\textwidth]{stable_value_estimation/cartpole_summary}
    \vspace{-3ex}
    \caption{
    Epistemic value estimation on DMC Cartpole Swingup across prior complexities $\ell$.
    \textit{Left:} OOD-detection AUROC of the predicted uncertainty $\sigma_Q$.
    \textit{Middle:} Pearson correlation between predicted uncertainty and value-estimation error.
    \textit{Right:} Predicted initial-state value over training, compared with the Monte Carlo reference.
    TUD improves both uncertainty metrics across length scales and tracks the Monte Carlo value more closely.
    }
    \label{fig:value_est_cartpole}
    \vspace{-3ex}
\end{figure}

\textbf{Temporal Uncertainty Difference Learning\;}
We propose to disentangle these signals architecturally and in the loss.
We split $f_\theta$ into a base network $b_{\theta_b}(s,a)$ carrying the mean value, a corrector $c_{\theta_c}(s,a,z)$ that is trained to cancel the prior on visited states, and a residual bootstrap network $rb_{\theta_r}(s,a,z)$ that propagates future residuals: $Q_\theta(s,a,z) = b_{\theta_b}(s,a) + (rb_{\theta_r} + c_{\theta_c} + g)(s,a,z)$.
Instead of the joint target in Equation~\ref{eq:joint}, we introduce and minimize the Temporal Uncertainty Difference (TUD) loss $\mathcal{L}^\text{TUD}$,
\begin{align}
\mathcal{L}^\text{TD}(\theta)
=\mathbb{E}_{\mathcal{B},p_z,\pi}\big[&\big(r+\gamma(b_{\theta_b} + rb_{\theta_r} + c_{\theta_c} + g)(s',a',z)\label{eq:joint}\\
-&(b_{\theta_b} +rb_{\theta_r} + c_{\theta_c} + g)(s,a,z))\big)^2]\\
\leq 3\mathbb{E}_{\mathcal{B},p_z,\pi}\big[
&\big(r + \gamma b_{\bar{\theta}_b}(s',a') - b_{\theta_b}(s,a)\big)^2 \label{eq:base}\\
+ &\big(g(s,a,z) + c_{\theta_c}(s,a,z)\big)^2 \label{eq:corrector}\\
+ &\big(\gamma (g + c_{\bar{\theta}_c} + rb_{\bar{\theta}_r})(s',a',z) - rb_{\theta_r}(s,a,z)\big)^2 \label{eq:residual_bootstrap}\big]=3\mathcal{L}^\text{TUD}(\theta).
\end{align}
Here, the upper bound holds, because the TUD loss discards the cross-terms between the different components (see Appendix~\ref{dis:tud_learning} for a full derivation).
It is exactly these cross terms that allow components to compensate for each other's errors and thereby allow the prior to leak into the reward signal.
Each term now isolates one signal: the base network is trained by standard TD-learning on the reward alone (Equation~\ref{eq:base}), the corrector by regression towards $-g$ (Equation~\ref{eq:corrector}), and the residual bootstrap network on the future residual $(g + c_{\bar{\theta}_c})(s',a',z)$ (Equation~\ref{eq:residual_bootstrap}).


\textbf{TUD learning stabilizes value estimation.\;}
To isolate the effect of TUD learning, we consider policy evaluation: we collect data from a policy, taken from an intermediate SAC checkpoint on DMC cartpole swingup, and learn its epistemic value function.
We consider a set of fixed states in the environment and label each of them in-distribution or out-of-distribution according to the policy's state-visitation density, estimated by kernel density estimation.
We evaluate whether uncertainty is high on OOD states and whether its magnitude tracks the value-estimation error.
Figure~\ref{fig:value_est_cartpole} shows that TUD learning achieves better calibration and better value fit across prior complexities.
This supports our motivation: under TD learning, the prior leaks into the Bellman target as reward noise, while TUD learning prevents this by bootstrapping the reward and the prior residual separately.
We provide implementation details and further experimental results in Appendix~\ref{app:td_with_enns} and ~\ref{app:ext_res_value_est}, respectively.

\textbf{Optimistic Value Estimation\;}
The TUD loss propagates signed residuals through the residual bootstrap network.
While this preserves the full distribution over epistemic value functions, we find that propagating residual magnitudes yields a more stable and effective exploration signal.
We therefore introduce TUD-OPT, which propagates the absolute residual,
\begin{align}
    \mathcal{L}^{\text{TUD-OPT}}(\theta_{r})=\mathbb{E}_{\mathcal{B},p_z,\pi}\big[\big(\gamma(|g+c|+rb_{\theta_r}(s',a',z)-rb_{\theta_r}(s,a,z)\big)^2\big]
\end{align}
This minimal change turns the residual bootstrap into a direct optimistic estimate of reachable novelty.
As we show in Figure~\ref{fig:ablation_components_coverage}, this yields a substantially stronger exploration signal and consistently improves exploration performance.

\subsection{Optimization: Maintaining Plasticity in Optimistic Exploration}
\begin{wrapfigure}{r}{0.6\textwidth}
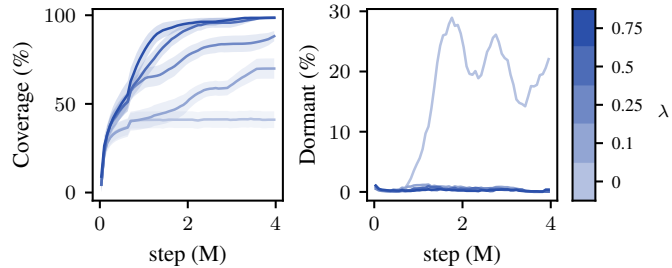

    \centering
    \vspace{-3ex}
    \incplt[0.6\textwidth]{plasticity_regularization/reg_summary}
    \vspace{-3ex}
    \caption{
    State coverage and the fraction of dormant neurons~\citep{sokar2023dormantneuronphenomenondeep} in the U-shaped PointMaze environment across soft-reset strengths.
    Increasing the strength of soft resets increases exploration performance and avoids loss of plasticity.}
    \label{fig:plasticity}
    \vspace{-2ex}
\end{wrapfigure}
Optimistic exploration introduces a highly non-stationary objective: as the agent collects informative data, the epistemic uncertainty driving exploration is reduced.
Under TUD learning, the corrector cancels the prior on visited state--action pairs, continuously changing the residual bootstrap target and, consequently, the exploration actor's objective.
Deep networks can lose the ability to adapt under such non-stationarity, a phenomenon known as loss of plasticity \citep{nikishin2022primacybiasdeepreinforcement, lyle2023understandingplasticityneuralnetworks, Dohare2024, sokar2023dormantneuronphenomenondeep}.
For exploration, this can cause the policy to act on a stale objective and stop reaching informative states.

\textbf{Soft Resets\;}
We counteract this loss of plasticity with soft resets \citep{nikishin2022primacybiasdeepreinforcement}, which periodically interpolate parameters toward a new random initialization,
\begin{align}
\theta_{t+1} \leftarrow (1-\lambda_t)\tilde{\theta}_{t+1} + \lambda_t\theta_0, \qquad \theta_0 \sim p_0(\theta),
\label{eq:soft-reset}
\end{align}
where $\tilde{\theta}_{t+1}$ denotes the parameters after the standard optimization step.
We set $\lambda_t=\lambda$ at fixed intervals and $\lambda_t=0$ otherwise, so that $\lambda$ directly controls the reset strength.
We apply soft resets only to the residual bootstrap network and exploration actor, whose objectives change with the epistemic signal, while leaving the base critic unchanged.
This periodically restores adaptability in the exploration components without discarding the learned task value.

\textbf{Optimistic Exploration Requires Plasticity\;}
We evaluate the effect of plasticity regularization in a reward-free U-shaped PointMaze, where continued exploration requires the policy to repeatedly adapt as the exploration frontier moves.
Figure~\ref{fig:plasticity} shows that stronger soft resets substantially improve state coverage.
Without resets, the fraction of dormant neurons --- neurons with normalized activation below a threshold $\tau$ \citep{sokar2023dormantneuronphenomenondeep} --- grows over training and the policy eventually stalls, whereas resets suppress this degradation and allow exploration to continue.
These results indicate that maintaining plasticity is necessary for the policy to keep tracking the moving epistemic objective and sustain exploration.
We outline the exact experimental details and further results in Appendix~\ref{app:impl_details_plasticity_reg} and ~\ref{app:plasticity_reg_res}, respectively.

\begin{algorithm}[h]
\caption{DEVOTE training step}
\label{alg:devote}
\begin{algorithmic}[1]
\Require replay buffer $\mathcal{B}$, step size $\eta$, reset rate $\lambda$,
  UCB coefficient $\beta$, temperature $\alpha$, replay ratio $k$, number of indices $M$
\For{$t = 1 \dots k$}
  \State $(s,a,r,s') \sim \mathcal{B}$, \; $z \sim p_z$, \;
         $a'_e \sim \pi_{\phi_e}(s')$, \; $a'_o \sim \pi_{\phi_o}(s')$
  \State \textbf{Update epistemic critic}
  \State $\theta_b \leftarrow \theta_b - \eta \nabla_{\theta_b}
         \big(r + \gamma\, b_{\bar\theta_b}(s',a'_e) - b_{\theta_b}(s,a)\big)^2$
  \State $\theta_c \leftarrow \theta_c - \eta \nabla_{\theta_c}
         \big((c_{\theta_c} + g)(s,a,z)\big)^2$
  \State $\theta_r \leftarrow \theta_r - \eta \nabla_{\theta_r}
         \big(\gamma\,(|c_{\bar\theta_c} + g| + rb_{\bar\theta_r})(s',a'_o,z)
         - rb_{\theta_r}(s,a,z)\big)^2$
  \State $\bar\theta \leftarrow \textsc{Polyak}(\theta, \bar\theta)$
  \State \textbf{Update actors}
  \State $a_o \sim \pi_{\phi_o}(s)$, \; $a_e \sim \pi_{\phi_e}(s)$, \;
         $z_i \sim p_z$, $i \in [M]$
  \State $\hat\mu_{Q_\beta}(s,a_o) \leftarrow \frac{1}{M}\sum_{i=1}^{M}
         \big[\, b_{\theta_b}(s,a_o)
         + \beta\,(rb_{\theta_r} + |c_{\theta_c} + g|)(s,a_o,z_i) \,\big]$
  \State $\phi_e \leftarrow \phi_e + \eta \nabla_{\phi_e}
         \big(b_{\theta_b}(s,a_e) - \alpha \log \pi_{\phi_e}(a_e \mid s)\big)$
  \State $\phi_o \leftarrow \phi_o + \eta \nabla_{\phi_o}
         \big(\hat\mu_{Q_\beta}(s,a_o) - \alpha \log \pi_{\phi_o}(a_o \mid s)\big)$
  \State \textbf{Plasticity regularization}
  \State $\psi \leftarrow (1-\lambda)\psi + \lambda \psi_0$, \;
         $\psi_0 \sim p_0(\psi)$, \; for $\psi \in \{\theta_r, \phi_o\}$
\EndFor
\end{algorithmic}
\end{algorithm}

\section{DEVOTE}\label{sec:devote}
As the previous section established, current applications of deep epistemic value functions are limited by challenges in estimating, propagating and optimizing uncertainty.
We introduce DEVOTE, a model-free exploration algorithm that enables scalable, principled exploration by preserving local epistemic uncertainty, propagating it over long horizons, and stably optimizing the non-stationary objective.
The residual $r_\theta(s,a,z)=|(c_{\theta_c}+g)(s,a,z)|$ acts as a local novelty signal.
It is small on well-covered state-action pairs and remains large where the agent has not gathered sufficient data.
The residual bootstrap critic $rb_{\theta_r}$ propagates this novelty under the exploration policy, converting local epistemic uncertainty into a temporally extended estimate of reachable novelty.
DEVOTE maintains an exploitation actor $\pi_{\phi_e}$, which maximizes the current mean value estimate $b_{\theta_b}$, and an exploration actor $\pi_{\phi_o}$, which maximizes the optimistic value
\begin{equation}
Q^\beta_\theta(s,a) = \mathbb{E}_{z\sim p_z} \left[ b_{\theta_b}(s,a) + \beta\big(rb_{\theta_r}+r_\theta\big)(s,a,z) \right]. \label{eq:devote_q}
\end{equation}
The base critic must bootstrap under the exploitation actor, since the exploration actor would distort the mean value estimate.
The exploration actor is used for all environment interaction and converges to the exploitation actor as the optimal policy is identified.
This is the analogue of principled upper-confidence-bound exploration, as the base critic captures expected task value, while the epistemic term provides an optimism bonus for actions that may lead to informative regions \citep{Srinivas_2012}.
The base critic bootstraps with actions from $\pi_{\phi_e}$, while the residual bootstrap critic bootstraps with actions from $\pi_{\phi_o}$, so the epistemic value reflects novelty the exploration policy can reach.
We normalize $b_{\theta_b}$ with robust percentiles \citep{hafner2024masteringdiversedomainsworld} to make its scale comparable to the epistemic term and fix $\beta=1$ across environments.
Soft resets are applied to $rb_{\theta_r}$ and $\pi_{\phi_o}$ to maintain adaptation to the non-stationary exploration objective.
Algorithm~\ref{alg:devote} summarizes the complete training step.


\section{Results}\label{sec:results}

We evaluate DEVOTE across five environments spanning pure exploration and high-dimensional continuous control, and organize the results around three main insights.
For all experiments, we report the mean across five seeds together with the standard error.
Additional implementation details and results are provided in Appendices~\ref{app:hyperparams}, and ~\ref{app:main_res}, respectively.

\textbf{Experimental Settings\;}
We consider two experimental settings.
\begin{itemize}
    \item \textbf{Pure exploration.}
    In the reward-free setting, novelty alone drives behavior, isolating how effectively a method identifies and reaches novel states.
    We evaluate DEVOTE and the baselines on DeepSea \citep{osband_behaviour_2020} ($|\mathcal{S}|=50^2$, $|\mathcal{A}|=2$, 500k steps), a custom spiral PointMaze ($\dim(\mathcal{S})=\dim(\mathcal{A})=2$, 5M steps), and DMC Cheetah \citep{tassa2018deepmindcontrolsuite} ($\dim(\mathcal{S})=17$, $\dim(\mathcal{A})=6$, 5M steps).
    We report state coverage, defined as the fraction of discretized state bins visited at least once ($x$--$y$ position for PointMaze and $x$--$y$ velocity for Cheetah).

    \item \textbf{High-dimensional control.}
    To test whether directed exploration remains effective when it must be balanced against task reward, we evaluate DEVOTE and the baselines on two high-dimensional control tasks with shaped rewards and sparse goals.
    In AntMaze \citep{freeman_brax_2021} ($\dim(\mathcal{S})=31$, $\dim(\mathcal{A})=8$), a goal-directed velocity reward guides the ant, while a wall blocks the direct path.
    In PickAndPlace \citep{zakka_mujoco_2025} ($\dim(\mathcal{S})=27$, $\dim(\mathcal{A})=8$), a lifting reward encourages raising the object without providing information about the target location.
    We run the tasks for 40M steps and report normalized return.
\end{itemize}

\begin{figure}[t]
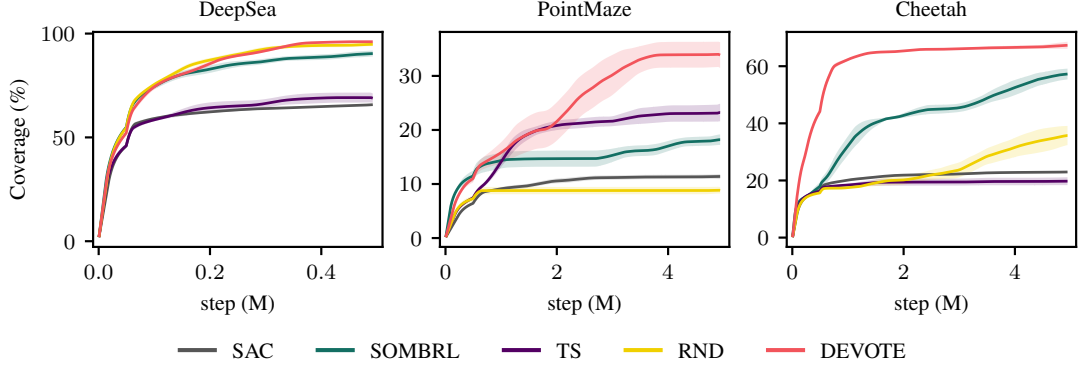

    \centering
    \vspace{-3ex}
    \incplt[\textwidth]{figures_8pt/pure_exploration/main_coverage}
    \vspace{-3ex}
    \caption{State coverage over training for DEVOTE and the baselines in the pure-exploration environments. DEVOTE consistently drives exploration toward novel states, matching or outperforming all baselines across tasks.}
    \label{fig:main_coverage}
    \vspace{-3ex}
\end{figure}

\textbf{Baselines\;}
We compare DEVOTE against the following baselines, with all methods sharing the same SAC backbone and common hyperparameters.
\begin{enumerate}
    \item \textbf{SAC}: Soft Actor-Critic \citep{haarnoja_soft_2018} serves as the common backbone and explores only through policy entropy.
    \item \textbf{Random Network Distillation (RND)}: RND \citep{burda2018explorationrandomnetworkdistillation} derives an intrinsic reward from the prediction error of a network trained to match a fixed randomly initialized target network.
    \item \textbf{Thompson Sampling (TS)}: Following randomized prior functions \citep{osband2018randomizedpriorfunctionsdeep}, we train an ensemble of independent SAC actor-critics on bootstrapped data, each augmented with a fixed randomly initialized MLP prior, and sample one actor per episode for data collection.
    \item \textbf{SOMBRL}: Scalable Optimistic Model-Based RL \citep{sukhija2025sombrlscalableoptimisticmodelbased} is a model-based exploration method that derives exploration signals from epistemic uncertainty in a learned dynamics model.
    We adapt SOMBRL to train SAC on its intrinsic rewards rather than on model rollouts, with uncertainty estimated using a Dreamer-based dynamics model \citep{hafner2024masteringdiversedomainsworld}.
\end{enumerate}

\textbf{Insight 1: Robust deep epistemic value functions drive exploration consistently.\;}
We first evaluate DEVOTE in pure exploration, where task reward is removed so that performance depends entirely on the quality of the exploration signal.
Figure~\ref{fig:main_coverage} shows the state coverage of DEVOTE and the baselines across the three pure-exploration environments.
DEVOTE outperforms all baselines on PointMaze and Cheetah and matches the strongest baseline on DeepSea, whereas the baselines perform inconsistently across tasks.
SAC explores only through action dithering and achieves poor coverage throughout.
The prior-based baselines are highly environment dependent: RND improves over SAC on DeepSea and Cheetah but fails on PointMaze, while TS performs well on PointMaze but provides little benefit elsewhere.
SOMBRL improves over SAC in all three environments, making its model-based uncertainty signal more consistent than the other baselines in this comparison, but still falls short of DEVOTE on PointMaze and Cheetah.
These results support our central claim that deep epistemic value functions can drive exploration reliably when their uncertainty is calibrated, propagated stably over time, and optimized robustly.
Notably, DEVOTE achieves this while remaining fully model-free.

\begin{figure}[t]
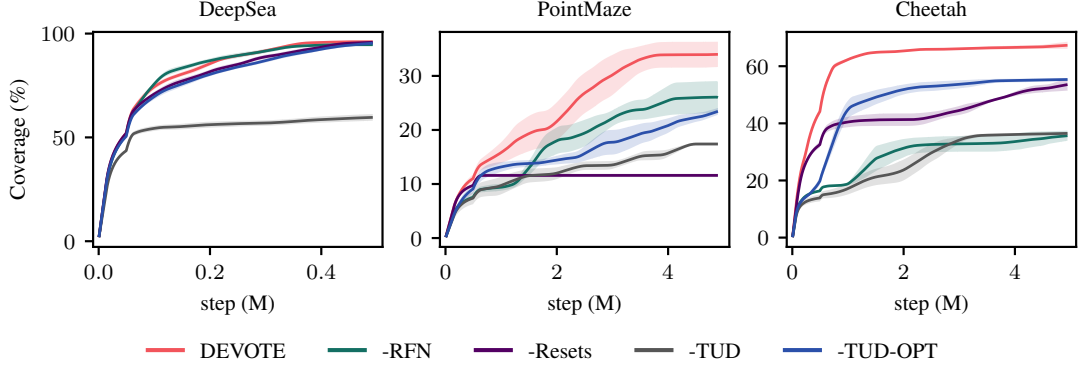

    \centering
    \vspace{-3ex}
    \incplt[\textwidth]{figures_8pt/pure_exploration/devote_ablation_coverage}
    \vspace{-3ex}
    \caption{State coverage over training when ablating the DEVOTE components introduced in Section~\ref{sec:limitations}.
    All components contribute in the continuous exploration tasks, with their relative importance depending on the structure of the environment.}
    \label{fig:ablation_components_coverage}
    \vspace{-3ex}
\end{figure}

\textbf{Insight 2: The DEVOTE components address complementary exploration failures.\;}
Having shown in Section~\ref{sec:limitations} that each component addresses its targeted failure mode in isolation, we now test whether these improvements translate into exploration performance.
In Figure~\ref{fig:ablation_components_coverage}, we remove each component in turn to isolate its contribution.
TUD learning improves coverage across all environments by training the corrector directly against the prior, sharpening the distinction between visited and novel regions.
On PointMaze and Cheetah, all components contribute, but their relative importance differs.
On DeepSea, the one-hot state encoding makes almost any random prior a useful novelty signal, reducing the importance of the RFN prior.
On PointMaze, soft resets matter most because the spiral continually shifts the exploration frontier, requiring the policy to remain plastic enough to track it.
On Cheetah, the main challenge is instead producing and propagating a useful novelty signal in a higher-dimensional state space, making the RFN prior and TUD(-OPT) learning more important while resets contribute less.
TUD-OPT further stabilizes the temporal propagation of novelty, improving performance on both PointMaze and Cheetah.
Overall, the RFN prior determines whether the agent obtains a meaningful local uncertainty signal, TUD learning determines whether that signal is learned reliably, TUD-OPT determines whether it is propagated over time, and soft resets determine whether the exploration policy can continue to adapt to and reach novel regions.

\begin{figure}[t]
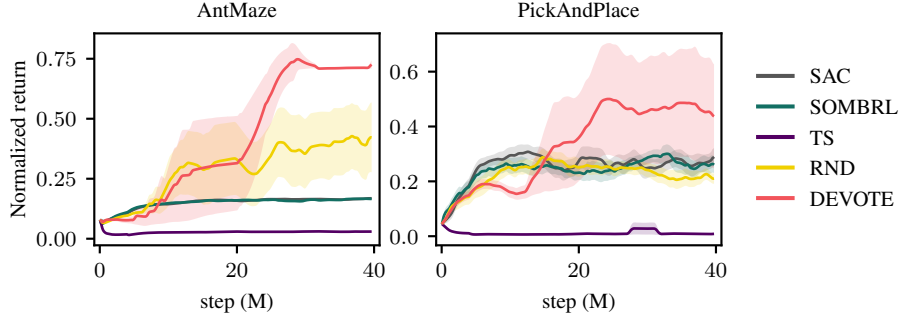

    \centering
    \vspace{-3ex}
    \incplt[0.84\textwidth]{figures_8pt/complex_control/main_normalized_return}
    \vspace{-7ex}
    \caption{Normalized return over training in the high-dimensional control environments. DEVOTE achieves the strongest task performance across both environments, indicating that its exploration signal can be combined effectively with extrinsic reward.}
    \label{fig:main_normalized_return}
    \vspace{-3ex}
\end{figure}

\textbf{Insight 3: DEVOTE effectively trades off exploration and exploitation.\;}
In Figure~\ref{fig:main_normalized_return}, we evaluate whether DEVOTE can balance epistemic exploration with task reward in the high-dimensional control environments.
In both AntMaze and PickAndPlace, the shaped reward teaches a useful skill but leads to a local optimum, so solving the task additionally requires exploration to discover the sparse-reward goal.
DEVOTE achieves the highest return in both environments and reaches the sparse goal earlier than the baselines.
This shows that the epistemic exploration signal complements the extrinsic objective: DEVOTE exploits the shaped reward to acquire useful behavior while continuing to explore beyond the resulting local optimum.

\section{Related Work}

\textbf{Sequential Decision Making\;}
Theoretical work on sequential decision making motivates uncertainty-directed exploration through optimism, posterior sampling, and information-directed strategies \citep{auer2002finite, thompson1933likelihood, Srinivas_2012, russo_learning_2017, kirschner_information_2018}.
In reinforcement learning, model-based approaches instantiate these principles by maintaining uncertainty over the unknown MDP and planning optimistically or under sampled models \citep{jaksch_near-optimal_2010, osband2013moreefficientreinforcementlearning, azar_minimax_2017, curi2020efficientmodelbasedreinforcementlearning}.
A complementary line places uncertainty directly on value functions, yielding optimistic and randomized value-based algorithms with strong guarantees \citep{osband2016generalizationexplorationrandomizedvalue, russo_worst-case_2019, wang_reinforcement_2020, ishfaq_randomized_2021, zanette_frequentist_2023}.
Related work has further shown that model-based uncertainty itself can be propagated recursively through a Bellman-style equation, providing a direct mechanism for temporally extended exploration \citep{odonoghue_uncertainty_2018, janz_successor_2019, luis_model-based_2023}.
DEVOTE follows the value-based perspective and focuses on scaling these principles.

\textbf{Deep Reinforcement Learning\;}
Modern deep RL scales value learning and policy optimization to high-dimensional problems using neural function approximation \citep{Mnih2015, lillicrap_continuous_2015, schulman_proximal_2017, fujimoto_addressing_2018, haarnoja_soft_2018}.
Standard agents typically rely on largely undirected exploration through $\epsilon$-greedy action selection, stochastic policies, or injected noise \citep{mnih_asynchronous_2016, plappert_parameter_2018, fortunato_noisy_2019}.
A large body of work instead introduces directed exploration through pseudo-counts and density estimation \citep{bellemare_unifying_2016, ostrovski_count-based_2017}, prediction error and curiosity \citep{pathak_curiosity-driven_2017, burda2018explorationrandomnetworkdistillation}, model disagreement \citep{pathak_self-supervised_2019, sekar_planning_2020, sukhija2025sombrlscalableoptimisticmodelbased}, and uncertainty-aware value functions \citep{osband2016deepexplorationbootstrappeddqn, osband2018randomizedpriorfunctionsdeep, ciosek_better_2019, dwaracherla_hypermodels_2020, lee_sunrise_2020}.
Directed exploration can substantially improve performance on hard-exploration problems \citep{ecoffet_go-explore_2021, sukhija_maxinforl_2025, diaz-bone_discover_2025}.
At the same time, deep RL remains difficult to optimize robustly at scale due to bootstrapping, non-stationarity, and loss of plasticity \citep{hasselt_deep_2018, nikishin2022primacybiasdeepreinforcement, lyle2023understandingplasticityneuralnetworks, nauman_bigger_2024}.
DEVOTE addresses these challenges by improving how epistemic uncertainty is represented, propagated, and optimized in deep epistemic value functions.

\section{Conclusion}
In this work, we studied how epistemic uncertainty must be represented, propagated, and optimized to make deep epistemic value functions a reliable mechanism for optimistic exploration.
Our empirical analysis shows that uncertainty must generalize meaningfully beyond observed data, propagate stably over long horizons, and remain actionable as the exploration objective changes with experience.
These findings motivate DEVOTE, which combines structured function-space priors, Temporal Uncertainty Difference learning, and plasticity regularization to address these three challenges jointly.
Across reward-free exploration and challenging continuous-control tasks, DEVOTE yields more reliable exploration and stronger task performance than competitive model-free and model-based baselines.\par

Our study is primarily empirical and focuses on state-based continuous-control settings. 
We do not establish theoretical guarantees on uncertainty calibration or exploration efficiency, and our current prior construction relies on relatively simple smoothness assumptions that may become limiting in higher-dimensional or representation-learning settings. 
Extending these ideas to richer learned representations, visual observations, and stronger theoretical characterizations is an important direction for future work. 
More broadly, our results suggest that scaling principled exploration requires not only the right algorithmic objective, but also a deep empirical understanding of how that objective interacts with neural function approximation, and the co-development of methods and implementations that remain robust in practice.

\newpage

\ifarxiv
\fi
\ifarxiv
  \subsubsection*{Acknowledgments}
    We would like to thank Maximilian Seelinger, Manuel Wendl and Klemens Iten for feedback on early versions of the paper. 
    Leander Diaz-Bone and Marco Bagatella are supported by the Max Planck ETH Center for Learning Systems.
    Jonas Hübotter was supported by the Swiss National Science Foundation under NCCR Automation, grant agreement 51NF40 180545.
\fi

\bibliography{iclr2027_conference}
\bibliographystyle{iclr2027_conference}
\newpage
\appendix
\section{Extended Background}\label{app:ext_back}

\subsection{Model-free Reinforcement Learning}\label{app:ext_back_mf_rl}

Model-free reinforcement learning estimates the value of a policy directly from experience, without maintaining an explicit model of the environment dynamics.
A value function summarizes the long-term consequences of an action in a quantity that can guide future decisions and can be learned from transitions $(s,a,r,s')$ using the Bellman equation \citep{sutton2018reinforcement}.
For a policy $\pi$, temporal-difference learning fits $Q_\theta$ to a bootstrapped target,
\begin{equation}
    \mathcal L^{\text{TD}}(\theta)
    =
    \mathbb E_{(s,a,r,s')\sim\mathcal B,\,
    a'\sim\pi(\cdot\mid s')}
    \left[
    \left(
    r+\gamma Q_{\bar\theta}(s',a')-Q_\theta(s,a)
    \right)^2
    \right].
\end{equation}
The replay buffer $\mathcal B$ stores past transitions, and $\bar\theta$ is a slowly updated copy of the critic parameters that is held fixed when differentiating the loss.
Because the next action is drawn from the policy being evaluated, the transition itself may have been collected by an earlier policy, enabling off-policy reuse of experience.

\paragraph{Policy improvement.}
A critic estimates the value of actions under a policy; policy improvement uses this estimate to choose better actions.
In discrete action spaces, Q-learning performs this improvement directly by maximizing over next actions in the Bellman target \citep{watkins_q-learning_1992,Mnih2015}.
In continuous action spaces, repeatedly solving this maximization can be expensive, so an actor $\pi_\phi$ learns to produce high-value actions instead.
Off-policy actor-critic methods alternate critic updates with actor updates that maximize
$\mathbb E_{s\sim\mathcal B,\,a\sim\pi_\phi}[Q_\theta(s,a)]$.
DDPG uses a deterministic actor, while TD3 additionally uses the smaller of two critic estimates in its target to reduce overestimation \citep{lillicrap_continuous_2015,fujimoto_addressing_2018}.

\paragraph{Soft Actor-Critic.}
SAC makes policy improvement entropy-regularized, favoring policies that assign probability to high-value actions while retaining stochasticity \citep{haarnoja_soft_2018}.
Its actor minimizes
\begin{equation}
    \mathcal L_\pi(\phi)
    =
    -\mathbb E_{s\sim\mathcal B,\,
    a\sim\pi_\phi(\cdot\mid s)}
    \left[
    Q_\theta(s,a)
    + \alpha \mathcal H[\pi_\phi(\cdot\mid s)]
    \right],
\end{equation}
where the temperature $\alpha$ controls the reward--entropy trade-off and can be learned against a target entropy.
Standard SAC also includes future entropy in the critic target.
Our SAC-based implementation retains entropy-regularized actor updates, but uses reward-only critic targets and omits the twin-critic minimum.
DEVOTE uses this separation to learn task value and an epistemic exploration objective with distinct critics and actors.
Appendix~\ref{app:impl_detail} gives further implementation details.

\subsection{Uncertainty Quantification}
Given noisy observations $\mathcal D_n=\{(x_i,y_i)\}_{i=1}^n$ of an unknown function $f$, uncertainty quantification aims to represent a belief $q_n\in\Delta(\mathbb{R}^{\mathcal X})$ over functions that remain plausible given the observed data.
The standard Bayesian approach formalizes this by placing a prior distribution $q_0$ over functions and conditioning it on $\mathcal D_n$ to obtain the posterior $q_n$.
Predictions can then be summarized by the belief mean $\mu_n(x)=\mathbb E_{f\sim q_n}[f(x)]$, while its variance $\sigma_n^2(x)=\operatorname{Var}_{f\sim q_n}[f(x)]$ quantifies the remaining epistemic uncertainty.
Epistemic uncertainty reflects limited knowledge about $f$ and can decrease as more data is observed, whereas aleatoric uncertainty captures irreducible variability in the observations \citep{hullermeier_aleatoric_2021, krause_probabilistic_2025}.
For simplicity, we consider regression with $y_i=f(x_i)+\epsilon_i$, where $\epsilon_i\sim\mathcal N(0,\sigma_\epsilon^2)$ are independent.


\paragraph{Gaussian processes.}
Gaussian processes provide a canonical model for uncertainty quantification in function space, with exact Bayesian posteriors under standard regression assumptions and a well-developed theoretical foundation.
A Gaussian process (GP) is a distribution over functions for which every finite collection of function values is jointly Gaussian \citep{rasmussen2006gaussian}.
A prior $f\sim\mathcal{GP}(\mu_0,k_0)$ is specified by a mean function and a positive-semidefinite covariance kernel.
Under Gaussian observation noise, conditioning on $\mathcal D_n$ gives a closed-form posterior with
\begin{align}
\mu_n(x) &= \mu_0(x) + k_{x,X}^\top (K_X+\sigma_\epsilon^2 I)^{-1}(y-\mu_0(X)),\\
k_n(x,x') &= k_0(x,x') - k_{x,X}^\top (K_X+\sigma_\epsilon^2 I)^{-1}k_{x',X}.
\end{align}
Here $(K_X)_{ij}=k_0(x_i,x_j)$ and $(k_{x,X})_i=k_0(x,x_i)$, and the posterior epistemic variance is $\sigma_n^2(x)=k_n(x,x)$.
The kernel encodes assumptions about how observations constrain the function elsewhere.
For example, the linear kernel $k(x,x')=\phi(x)^\top\phi(x')$ specifies a feature representation, while the RBF kernel $k_{\mathrm{RBF}}(x,x')=\exp\left(-\|x-x'\|_2^2/(2\ell^2)\right)$ encodes smoothness at length scale $\ell$.
This explicit function-space structure makes GPs particularly attractive for calibrated uncertainty estimation and theoretical analysis.
Their main limitation in our setting, however, is that standard GPs rely on a fixed kernel or feature representation, and therefore do not naturally learn rich task-dependent representations from high-dimensional data.

\paragraph{Random Fourier features.}
The $\mathcal{O}(n^3)$ cost of exact GP inference motivates finite-dimensional approximations.
Random Fourier features (RFFs) \citep{rahimi2007randomfourierfeatures, krause_probabilistic_2025} approximate a stationary kernel $k_0(x,x')=k_0(x-x')$ by a finite inner product of randomized feature maps, reducing GP regression to Bayesian linear regression in a feature space of dimension $m\ll n$.
The construction relies on Bochner's theorem: for any continuous shift-invariant kernel with $k_0(0)=1$, there exists a probability density $p_\omega$ such that
\begin{align}
k_0(x-x') &= \mathbb E_{\omega\sim p_\omega}\left[\cos\bigl(\omega^\top(x-x')\bigr)\right] \\
&= 2\,\mathbb E_{\omega\sim p_\omega,\,b\sim\mathcal U([0,2\pi])}\left[\cos(\omega^\top x+b)\cos(\omega^\top x'+b)\right].
\end{align}
For the RBF kernel $k_{\mathrm{RBF}}(x,x';\ell)$, the spectral density is $p_\omega=\mathcal N(0,\ell^{-2}I)$, where $\ell$ is the kernel length scale.
Drawing $m$ i.i.d. samples $\{(\omega_i,b_i)\}_{i=1}^m$ and defining
\begin{equation}
\varphi_m(x)=\sqrt{\frac{2}{m}}\bigl(\cos(\omega_1^\top x+b_1),\ldots,\cos(\omega_m^\top x+b_m)\bigr)^\top
\end{equation}
yields the Monte Carlo approximation $k_0(x,x')\approx\varphi_m(x)^\top\varphi_m(x')$.
Under this approximation, representing $f(x)=\varphi_m(x)^\top w$ with a Gaussian prior over $w$ reduces approximate GP inference to Bayesian linear regression with a Gaussian posterior available in closed form.
The fixed feature representation avoids the ever-growing kernel matrix of exact GP inference, but finite $m$ can degrade uncertainty estimates.
Because the model expresses $f$ through only $m$ basis functions, its posterior uncertainty is governed by the geometry of this fixed feature space, which can lead to \emph{variance starvation}: systematic underestimation of uncertainty away from the observed data \citep{wang_batched_2018}.\par
We note that the random Fourier feature construction can equivalently be used to define a single-hidden-layer random network.
We refer to these networks as Randomized Fourier Networks (RFNs),
\begin{equation}
g_{\mathrm{rfn}}(x,z)=\sqrt{\frac{2}{m}}\,\mathbf{1}^\top\cos(\Omega_z x+b_z),
\end{equation}
where the cosine is applied elementwise and the rows $\omega_{z,i}$ of $\Omega_z$ and biases $b_{z,i}$ are sampled as in the RFF construction above.
We use these RFNs as the fixed prior $g$ in our ENNs.


\subsection{Random Network Distillation}
Developing DEVOTE from the perspective of deep epistemic value functions also sheds light on the empirical success of Random Network Distillation (RND) \citep{burda2018explorationrandomnetworkdistillation}.
RND fits a predictor to a fixed random target network and uses the remaining prediction error as a local novelty signal,
\begin{equation}
r_{\mathrm{RND}}(s)=\lVert g(s)+c_\theta(s)\rVert^2.
\end{equation}
This is the same prior--corrector residual used by DEVOTE, and both methods propagate this residual through a learned value function to obtain a temporally extended exploration objective.
Viewed this way, RND can be understood as a simple instance of optimistic value learning in which a fixed random prior induces epistemic novelty and the corresponding value function predicts its future accumulation.

DEVOTE retains this basic mechanism, but replaces the single random prior with a distribution of RFN priors and regularizes the exploration components to preserve plasticity as the novelty signal evolves.
Our experiments show that both choices are empirically important for reliable exploration across environments (Section~\ref{sec:results}).
This perspective therefore helps explain why RND is already a strong exploration method, while also clarifying why seemingly small implementation choices in the prior and optimization can substantially affect the quality of the resulting epistemic value function.
\newpage
\section{Extended Discussion}

\subsection{TUD Learning} \label{dis:tud_learning}
In the following, we will expand on the discussion of Temporal Uncertainty Difference (TUD) Learning and its relationship to standard TD Learning for epistemic value functions.
We consider the following architecture for the $Q_\theta$ function:
\begin{align*}
    Q_\theta(s,a,z) = b_{\theta_b}(s,a) + (rb_{\theta_r} + c_{\theta_c}+g)(s,a,z),
\end{align*}
where $b_{\theta_b}$ is the base network not conditioned on $z$, $g$ is a fixed prior, $c_{\theta_c}$ is the corrector network and $rb_{\theta_r}$ is the residual bootstrap network.
We now consider the TD loss for this $Q$-function;
\begin{align*}
    \mathcal{L}^{\text{TD}}(\theta)
    = \mathbb{E}_{\mathcal{B},\pi,p_z}[&(r+\gamma Q_{\bar{\theta}}(s',a',z)-Q_\theta(s,a,z))^2]\\
    = \mathbb{E}_{\mathcal{B},\pi,p_z}[&(r+\gamma b_{\bar{\theta}_b}(s',a') + \gamma(rb_{\bar{\theta}_r} + c_{\bar{\theta}_c}+g)(s',a',z)\\
    &-b_{\theta_b}(s,a) - (rb_{\theta_r} + c_{\theta_c}+g)(s,a,z))^2] \\
    = \mathbb{E}_{\mathcal{B},\pi,p_z}[&(r+\gamma b_{\bar{\theta}_b}(s',a') - b_{\theta_b}(s,a)\\
    &- (c_{\theta_c} + g)(s,a,z)\\
    &+ \gamma (rb_{\bar{\theta}_r} + c_{\bar{\theta}_c}+g)(s',a',z) -rb_{\theta_r}(s,a,z))^2].  
\end{align*}
This arrangement groups the residual into three components, one per network.
We now expand the square;
\begin{align*}
     \mathcal{L}^{\text{TD}}(\theta)
     = \mathbb{E}_{\mathcal{B},\pi,p_z}[&(\underbrace{r+\gamma b_{\bar{\theta}_b}(s',a') - b_{\theta_b}(s,a)}_{\delta_b})^2\\
     &+ (\underbrace{(c_{\theta_c} + g)(s,a,z)}_{\delta_c})^2\\
     &+ (\underbrace{\gamma (rb_{\bar{\theta}_r} + c_{\bar{\theta}_c}+g)(s',a',z) -rb_{\theta_r}(s,a,z)}_{\delta_{rb}})^2]\\
     + 2\mathbb{E}_{\mathcal{B},\pi,p_z}[&-\delta_b\delta_c+\delta_b\delta_{rb}-\delta_c\delta_{rb}].
\end{align*}
This shows that the TD loss $\mathcal{L}^{\text{TD}}(\theta)$ is the sum of the TUD loss $\mathcal{L}^{\text{TUD}}(\theta)$ and three cross terms between the components, where the TUD loss is defined as:
\begin{align*}
    \mathcal{L}^{\text{TUD}}(\theta)
    = \mathbb{E}_{\mathcal{B},\pi,p_z}\big[&(r+\gamma b_{\bar{\theta}_b}(s',a') - b_{\theta_b}(s,a))^2\\
    &+ ((c_{\theta_c} + g)(s,a,z))^2\\
    &+ (\gamma (rb_{\bar{\theta}_r} + c_{\bar{\theta}_c}+g)(s',a',z) - rb_{\theta_r}(s,a,z))^2\big].
\end{align*}
\paragraph{Shared minimizers.}
We first argue that a minimizer of the TUD loss also minimizes the TD loss.
Since the target parameters $\bar{\theta}$ are held fixed, both objectives are least-squares regressions: the TD loss regresses $Q_\theta(s,a,z)$ onto $y = r + \gamma Q_{\bar{\theta}}(s',a',z)$, while the TUD loss regresses the base network onto $y_b = r + \gamma b_{\bar{\theta}_b}(s',a')$, the corrector onto $-g(s,a,z)$, and the residual bootstrap network onto $y_{rb} = \gamma (rb_{\bar{\theta}_r} + c_{\bar{\theta}_c} + g)(s',a',z)$.
Over an expressive function class, each regression is minimized by the conditional expectation of its target. Since $z$ is independent of the transition and the backup action, the TUD minimizers satisfy
$b^\star = \mathbb{E}[y_b \mid s,a]$, $c^\star = -g$ on the support of $\mathcal{B}\times p_z$, and $rb^\star = \mathbb{E}[y_{rb} \mid s,a,z]$. As $y_b + y_{rb} = y$, linearity of conditional expectation gives $(b^\star + rb^\star + c^\star + g)(s,a,z) = \mathbb{E}[y \mid s,a,z]$, the unique function-space minimizer of the TD loss. The converse fails: the TD loss constrains only the sum of the components, while the TUD loss breaks this degeneracy and assigns the epistemic component to the $z$-conditioned
channel.
Note that the cross terms in the expansion above do not vanish pointwise. The base and residual bootstrap residuals share the transition noise and are correlated in general. The minimizers coincide nonetheless because the argument operates on the three regressions directly.
\paragraph{Multi-iteration behavior.}
The decomposition also determines how the components behave across iterations of the update. Among the three regressions, only the corrector has a stationary and noiseless target: $g$ is fixed and $-g(s,a,z)$ depends only on the regression inputs, whereas the base and residual bootstrap networks chase
moving bootstrap targets. The corrector therefore drives $c_{\theta_c} + g$ to zero on the support of $\mathcal{B} \times p_z$ faster than the bootstrapped components, while off-support it can match $-g$ only through generalization, so $c_{\theta_c} + g$ retains variance across $z$ wherever the data does not constrain it. Unrolling the residual bootstrap recursion at its fixed point yields
\begin{align*}
    (rb^\star + c^\star + g)(s,a,z)
    = \mathbb{E}_{\pi}\Big[\sum_{t \geq 0} \gamma^t
      (c^\star + g)(s_t,a_t,z) \,\Big|\, s_0 = s, a_0 = a\Big],
\end{align*}
so the $z$-conditioned channel is the value function of the pseudo-reward $(c^\star + g)$ under the backup policy. Since this pseudo-reward vanishes on-support, the disagreement $\sigma_Q(s,a)=\text{Std}_{p_z}[Q_\theta(s,a,z)]$ is large only at poorly covered state-action pairs and at pairs whose successors lead toward them, with the contribution of future novelty discounted by $\gamma^t$.
TUD learning thus separates the epistemic signal into an instantaneous component, measured by the corrector residual, and a propagated component, accumulated by the residual bootstrap, thereby recovering the uncertainty Bellman-equation structure without a model.
\newpage
\section{Extended Related Work}\label{app:related_work}
\paragraph{Deep Uncertainty Quantification}
Bayesian neural networks approximate a posterior over weights through methods such as probabilistic backpropagation and dropout-based inference \citep{hernandezlobato2015probabilisticbackpropagationscalablelearning,gal_dropout_2016}, while stochastic weight averaging estimates a Gaussian approximation from optimization trajectories \citep{maddox_simple_2019}.
Deep ensembles represent uncertainty through independently trained predictors \citep{lakshminarayanan2017simplescalablepredictiveuncertainty}; randomized priors and ENNs extend this approach with explicit function diversity beyond the training data \citep{osband2018randomizedpriorfunctionsdeep,osband2023epistemicneuralnetworks}.
Their calibration remains sensitive to distribution shift and to gaps in the observed inputs \citep{ovadia2019trustmodelsuncertaintyevaluating,foong2019inbetweenuncertaintybayesianneural}.
Our RFN construction addresses prior design in ENNs and we evaluate whether the resulting uncertainty identifies unfamiliar inputs and tracks prediction error.

\paragraph{Plasticity and Value Learning}
Bootstrapping makes value learning sensitive to errors in its own targets, and repeated updates can impair adaptation to new data \citep{hasselt_deep_2018,nikishin2022primacybiasdeepreinforcement}.
Dormant-neuron recycling, parameter resets, and continual backpropagation address different manifestations of this loss of plasticity \citep{sokar2023dormantneuronphenomenondeep,nikishin2022primacybiasdeepreinforcement,Dohare2024}.
Normalization and critic ensembles also improve robustness when scaling networks or increasing replay \citep{ba_layer_2016,chen_randomized_2021,nauman_bigger_2024}.
DEVOTE's exploration objective adds a specific source of non-stationarity, as learning the prior changes both the residual-bootstrap target and the actions favored by the exploration actor.
We therefore apply soft resets to these two components while retaining the extrinsic base critic.

\newpage
\section{Implementation and Experimental Details}\label{app:impl_detail}
\subsection{Base Algorithm and Hyperparameters}\label{app:hyperparams}
The main experiments build on the DEVOTE agent from Algorithm~\ref{alg:devote}, operating on state observations.
Table~\ref{tab:base_hyperparams} lists the configuration shared across experiments.
The per-experiment sections state only deviations from it.
\definecolor{tablegroupgray}{gray}{0.45}

\providecommand{\appgroupheader}[1]{%
  \midrule
  \multicolumn{2}{l}{\color{tablegroupgray}#1}\\
  \midrule
}
\begin{table}[!htb]
\centering
\setlength{\tabcolsep}{8pt}
\begin{adjustbox}{max width=\linewidth}
\begin{tabular}{l r}
\toprule
\textbf{Hyperparameter} & \textbf{Value}\\
\midrule
\appgroupheader{Critic}
Critic architecture $b_{\theta_b}, rb_{\theta_r}$ & MLP (SiLU), $3 \times 1024$ + RMSNorm \\
Epistemic distribution $p_z$ & $\mathcal{U}([8])$ \\
Bootstrap probability & 0.5 \\
Polyak target network update & $0.005$ \\
\appgroupheader{Prior}
Prior, Corrector architecture $g,c_{\theta_c}$ & RFN, $1{,}024$ features \\
Prior length scale $\ell$ & 1 \\
Prior output activation & $\tanh$ \\
\appgroupheader{Optimization}
Optimizer & Adam \\
Learning rate $\eta$ & $3\times 10^{-4}$ \\
Critic weight decay & $10^{-2}$ \\
Actor weight decay & 0 \\
\appgroupheader{Actors}
Actor architecture $\pi_{\phi_e}, \pi_{\phi_o}$ & MLP (SiLU), $3 \times 192$ + RMSNorm \\
Policy distribution (continuous actions) & squashed normal \\
UCB coefficient $\beta$ & 1 \\
Initial temperature $\alpha$ & 1 \\
Target entropy scale & $-0.5\dim(\mathcal{A})$ \\
\appgroupheader{Data \& replay}
Discount $\gamma$ & 0.998 \\
Replay ratio $k$ & 1 \\
Batch size & $2{,}048$ \\
\appgroupheader{Soft resets}
Reset rate $\lambda$ & 0.5 \\
Reset interval (steps) & 250{,}000 \\
Reset scope & $\theta_r$, $\phi_o$ \\
\bottomrule
\end{tabular}
\end{adjustbox}
\caption{Shared agent hyperparameters, using the notation of Algorithm~\ref{alg:devote}. Experiment-specific overrides are stated in the text; prior length scales are given in Appendix~\ref{app:hyperparams}.}
\label{tab:base_hyperparams}
\end{table}

\paragraph{Exploration--exploitation trade-off.}
The extrinsic value inherits the scale of the environment return, whereas the epistemic bonus is determined only by intrinsic quantities.
To keep $\beta$ comparable across tasks, we rescale only the extrinsic contribution by a running estimate $S$ of its magnitude and train the exploration actor to maximize
\begin{equation}
\mathbb E_{s\sim\mathcal B,\,a\sim\pi_{\phi_o}}
\left[
\frac{b_{\theta_b}(s,a)}{S}
+\beta\,\mathbb E_{z\sim p_z}\left[(rb_{\theta_r}+|c_{\theta_c}+g|)(s,a,z)\right]
-\alpha\log\pi_{\phi_o}(a\mid s)
\right],
\end{equation}
where $S=\max(1,h-l)$ and $l$ and $h$ are exponential moving averages with update rate $0.01$ of the fifth and ninety-fifth percentiles of $b_{\theta_b}$ over replay batches, following the robust return normalization of \citet{hafner2024masteringdiversedomainsworld}.
We fix $\beta=1$ across environments.

\paragraph{Base actor-critic.}
DEVOTE uses a SAC-style off-policy actor-critic with several modifications motivated by recent findings.
We parameterize the extrinsic base critic as a categorical value distribution over symexp-spaced bins and train it using a two-hot likelihood \citep{hafner2024masteringdiversedomainsworld}, which we find to be more robust to overestimation.
In contrast, the residual bootstrap critic predicts scalar residuals and is trained using a mean-squared-error objective.
The residual carries small, intrinsic-scale values for which a direct regression head is better behaved than the exponentially spaced bins.
Following \citet{nauman_bigger_2024}, we omit SAC's twin-critic construction and clipped pessimistic target, as pessimistic targets can reduce performance when combined with categorical critics and normalized representations.
Both actors and critics apply RMS normalization throughout their hidden layers.
Finally, although the actor objectives remain entropy-regularized and the entropy temperature is learned, we omit the entropy term from the critics' Bellman targets.
This prevents the extrinsic value target from mixing environment rewards with an entropy contribution whose scale changes during training and improves performance in our experiments.

\paragraph{Prior selection.}\label{subsec:lengthscale_selection}
As shown in Section~\ref{app:main_res}, the RFN prior length scale can substantially affect exploration performance.
We select a single length scale for each environment from the fixed candidate set $\ell\in\{1,2.5,5,7.5\}$ based on a preliminary exploration-performance sweep.
The selected value is shared across all DEVOTE variants and reported in Table~\ref{tab:app-prior-lengthscales}.

\begin{table}[htbp]
\centering
\begin{tabular}{lr}
\toprule
Environment & Length scale $\ell$ \\
\midrule
DeepSea      & $1$   \\
PointMaze    & $1$   \\
Cheetah      & $7.5$ \\
AntMaze      & $7.5$ \\
PickAndPlace & $1$   \\
\bottomrule
\end{tabular}
\caption{\textbf{Environment-specific RFN length scales.}
For each environment, we select one value from the fixed candidate set $\ell\in\{1,2.5,5,7.5\}$ and share it across all DEVOTE variants.}
\label{tab:app-prior-lengthscales}
\end{table}

\subsection{ENN Calibration}\label{app:enn_cal_iml}
\paragraph{Data and splits.}
We use eight UCI regression datasets: Concrete, Energy, Kin8nm, Naval, Protein, Power, Wine, and Yacht.
Columns are rescaled using their first and ninety-ninth percentiles and clipped to $[0,1]$.
For each of ten gap splits \citep{foong2019inbetweenuncertaintybayesianneural}, points are sorted along one input dimension, the middle third is held out as OOD, and 10\% of the remaining points are reserved for ID evaluation.
For split $s$, the sorting dimension is $s\bmod D$, where $D$ is the input dimension.
The synthetic task samples a one-dimensional GP with an RBF kernel of length scale one on $[0,10]$, with observation noise $0.1$.
Training inputs form three clusters of 50 points, each within radius $0.1$ of a uniformly sampled anchor.
ID inputs follow the same clustered distribution and share the anchors, while 250 OOD inputs are sampled uniformly over the domain.
The exact GP posterior serves as a reference.

\paragraph{Models and training.}
To isolate uncertainty estimation, each prior-corrector pair is trained to minimize $(c+g)^2$ at the training inputs without using their labels.
A separate two-layer MLP of width 100, with weight decay $10^{-4}$, learns the predictive mean from the labels.
RFN priors have 1,024 features and length scales $\{0.25,0.5,1,2,4\}$.
MLP priors use ReLU activations, widths $\{50,100,200,400\}$, and one or two hidden layers; each corrector matches its prior's architecture.
The headline ENN-Ens-MLP comparison uses width 100 and depth two, selected from the architecture ablation.
Training uses Adam with learning rate $10^{-3}$, batch size 256, and 500 epochs.
The bootstrapped-ensemble baseline uses ten independently initialized mean predictors with independent Poisson(1) weights for each training point and member \citep{lakshminarayanan2017simplescalablepredictiveuncertainty}.


\paragraph{Metrics.}
We report the AUROC for separating ID points from \emph{misfit} OOD points, whose absolute prediction error exceeds the ninetieth percentile of ID errors.
We also report the Pearson correlation between predicted standard deviation and absolute prediction error on the joint evaluation set.
The noiseless sampled function is the reference on the GP task.
For UCI, a reference MLP trained on the full dataset supplies a denoised proxy; these correlations therefore measure agreement with that proxy, rather than with an observed noiseless ground truth.
Regression results are averaged over ten splits.

\subsection{TD-Learning with ENNs}\label{app:td_with_enns}
We freeze an intermediate SAC policy trained on task reward in DMC cartpole swingup or cheetah run, then train only the critics for two million environment steps.
The actor and temperature remain fixed.
We compare joint TD and disentangled TUD targets over RFN length scales $\{0.5,1,2.5,5,10\}$, averaging over three seeds.
The remaining training hyperparameters are unchanged from the policy-training configuration; only the value-function losses remain active.

At six anchor configurations per task, we evaluate 20 evenly spaced values of one velocity coordinate while holding the remaining coordinates fixed.
For cartpole, we sweep pole angular velocity over $[-12,12]$; for cheetah, we sweep torso forward velocity over $[-5,12]$.
The anchors range from commonly visited configurations to states far from the policy's visitation distribution; Tables~\ref{tab:app-cartpole-anchors} and~\ref{tab:app-cheetah-anchors} give their complete specifications.
Monte Carlo reference values average 16 rollouts of the frozen policy after the anchor action, with discount $0.99$ and up to 2,500 agent steps.
The episode limit is raised to 10,000 frames for this evaluation.

We estimate visitation density from 30,000 visited states using Gaussian kernel density estimation in per-coordinate standardized proprioceptive space, with Silverman's bandwidth rule.
The top and bottom terciles of log density define ID and OOD grid points; the middle tercile is excluded from the ID/OOD comparison.
At the final checkpoint, we report OOD AUROC and the correlation of predicted uncertainty with absolute Monte Carlo value error.

\begin{table}[htbp]
    \centering
    \begin{tabular}{lrrr}
        \toprule
        Cartpole anchor & Cart position $x$ & Pole angle $\theta$ & Cart velocity $v_x$\\
        \midrule
        Upright, balanced & 0 & 0 & 0\\
        Near top, offset & 0 & 0.6 & 0\\
        Mid-swing, climbing & 0.4 & 1.5 & 0.5\\
        Rail-pinned, hanging & $-1.85$ & $\pi$ & $-0.5$\\
        Racing upright & 0 & 0 & 6.0\\
        Rail-charging, hanging & 1.5 & $\pi$ & 4.0\\
        \bottomrule
    \end{tabular}
    \caption{Cartpole anchor configurations. Angles are in radians; anchor pole angular velocity and held action are zero. Pole angular velocity is swept over $[-12,12]$.}
    \label{tab:app-cartpole-anchors}
\end{table}
\begin{table}[htbp]
    \centering
    \begin{adjustbox}{max width=\linewidth}
    \begin{tabular}{lrrrrl}
        \toprule
        Cheetah anchor & $\Delta z$ & $\theta$ & $v_z$ & $\omega$ & Joint angles\\
        \midrule
        Crouched, low & $-0.25$ & 0 & 0 & 0 & 0, 0, 0, 0\\
        Stride phase & 0 & 0 & 0 & 0 & $0.5,-0.3,-0.4,0.2$\\
        Tilted run & 0 & 0.3 & 0 & 2.0 & 0, 0, 0, 0\\
        Hard landing & 0.35 & 0 & $-3.5$ & 0 & $0.5,-0.6,0.5,-0.6$\\
        Push-off, airborne & 0.2 & $-0.1$ & 2.0 & 0 & $0.6,-0.5,0.6,-0.5$\\
        Flipped & $-0.1$ & $\pi$ & 0 & 0 & 0, 0, 0, 0\\
        \bottomrule
    \end{tabular}
    \end{adjustbox}
    \caption{Cheetah anchors ordered by decreasing visitation density. Torso height offset $\Delta z$ and pitch $\theta$ are relative to the default state; $v_z$ and $\omega$ are vertical and pitch velocities. Joint angles list back thigh, back shin, front thigh, and front shin, in radians. Other held coordinates are zero. Forward velocity is swept over $[-5,12]$.}
    \label{tab:app-cheetah-anchors}
\end{table}

\subsection{Non-Stationarity in Epistemic Objectives}\label{app:impl_details_plasticity_reg}
The plasticity experiment uses a reward-free U-shaped PointMaze for 4M environment steps, requiring the agent to discover and traverse both arms.
Task reward is scaled to zero, and the agent propagates absolute residual magnitudes, as in DEVOTE.
We vary the soft-reset strength over $\{0,0.1,0.25,0.5,0.75\}$ with five seeds per setting.
Resets are applied every 50,000 steps to the exploration actor and residual bootstrap critic, overriding the default interval in Table~\ref{tab:base_hyperparams}.
Coverage is the fraction of visited cells in a uniform discretization of the maze.
We measure plasticity using the fraction of $\tau$-dormant neurons in the exploration actor \citep{sokar2023dormantneuronphenomenondeep}.
For neuron $i$ in a layer of width $d_k$, with activation $a_i(x)$ under input distribution $\mathcal D$, define
\begin{equation}
    s_i^k=\frac{\mathbb E_{x\sim\mathcal D}[|a_i(x)|]}
    {d_k^{-1}\sum_{j\in k}\mathbb E_{x\sim\mathcal D}[|a_j(x)|]}.
\end{equation}
A neuron is dormant when $s_i^k\leq\tau$; we use $\tau=0.05$ and report the fraction across the network.
Training curves show seed means and standard errors, with moving-average smoothing.

\subsection{Exploration Baselines and Reporting}
All exploration methods use the same SAC backbone and shared hyperparameters in Table~\ref{tab:base_hyperparams}.
SAC explores through policy entropy alone.
RND is implemented on top of SAC and uses the prediction error against a fixed, randomly initialized ReLU target network as intrinsic reward \citep{burda2018explorationrandomnetworkdistillation}.
For Thompson sampling, we use independent SAC actor-critics trained on bootstrapped data with fixed ReLU priors, sampling one actor per episode for data collection \citep{osband2018randomizedpriorfunctionsdeep}.
For SOMBRL, we use a Dreamer-based dynamics model to derive intrinsic rewards and train SAC on those rewards, instead of training a policy on model rollouts as in the original method \citep{sukhija2025sombrlscalableoptimisticmodelbased,hafner2024masteringdiversedomainsworld}.

The pure-exploration and high-dimensional-control comparisons report means over five seeds with standard errors.
We aggregate each seed separately within each environment. 
We divide the training budget into $500$ equally spaced bins, average all observations from a seed within each bin, and linearly interpolate internal empty bins between observed values. 
We then compute the mean and standard error across seeds and apply a causal $50$-bin moving average to both. 
Summary panels average the resulting environment-level curves with equal weight, and curves end at the last point for which every compared method has a defined mean and standard error. 
These seeds are separate from the ten regression splits and three policy-evaluation seeds above.

\begin{figure}[t]
\centering

\setlength{\tabcolsep}{0pt}
\setlength{\arrayrulewidth}{0.8pt}
\renewcommand{\arraystretch}{0}

\begin{tabular}{|@{}c@{}|@{}c@{}|@{}c@{}|}
\hline
\includegraphics[height=0.24\linewidth]{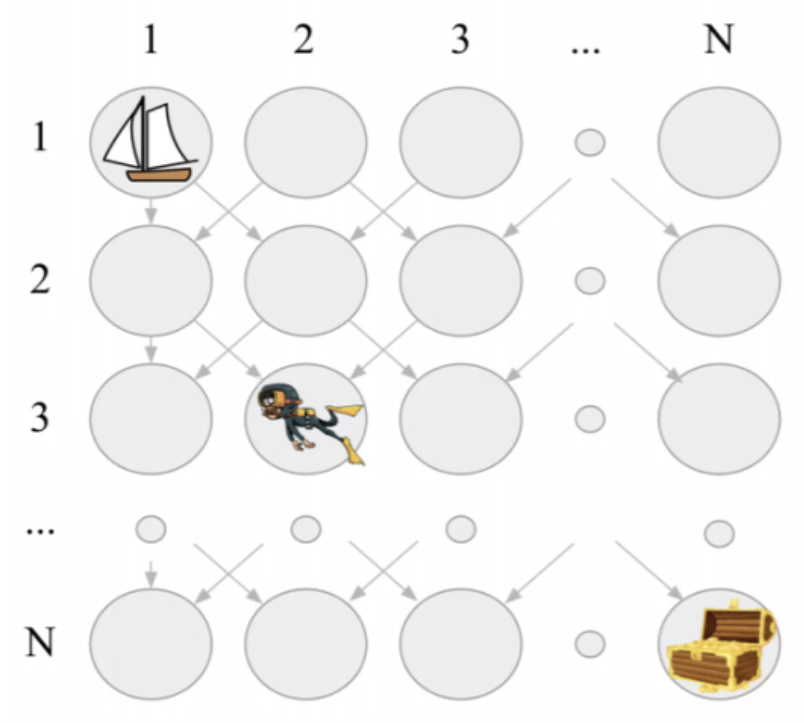} &
\includegraphics[height=0.24\linewidth]{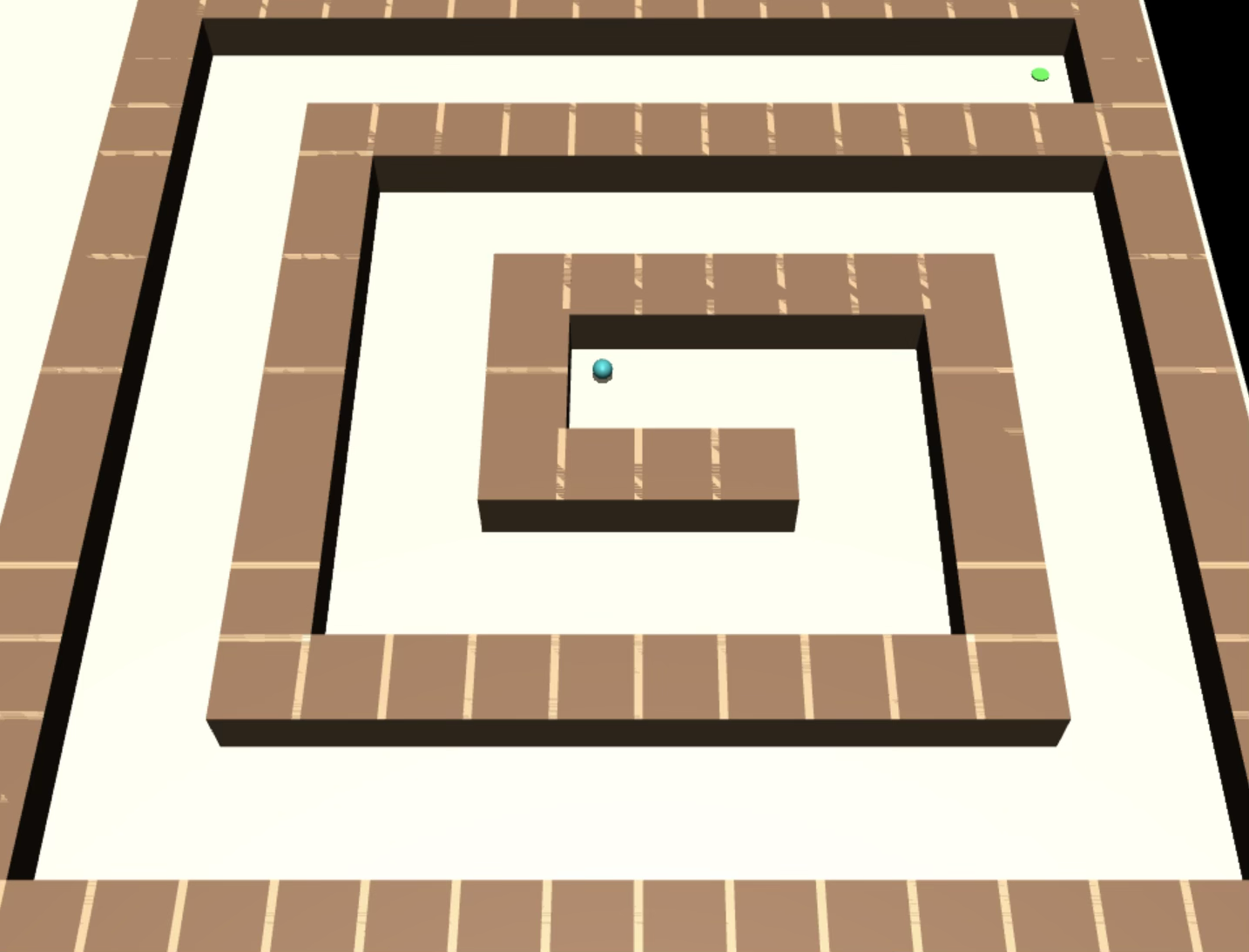} &
\includegraphics[height=0.24\linewidth]{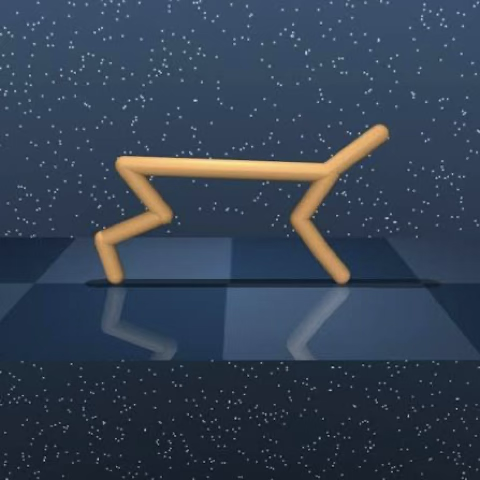} \\
\hline
\end{tabular}

\vspace{-0.8pt}

\begin{tabular}{|@{}c@{}|@{}c@{}|}
\includegraphics[height=0.353\linewidth]{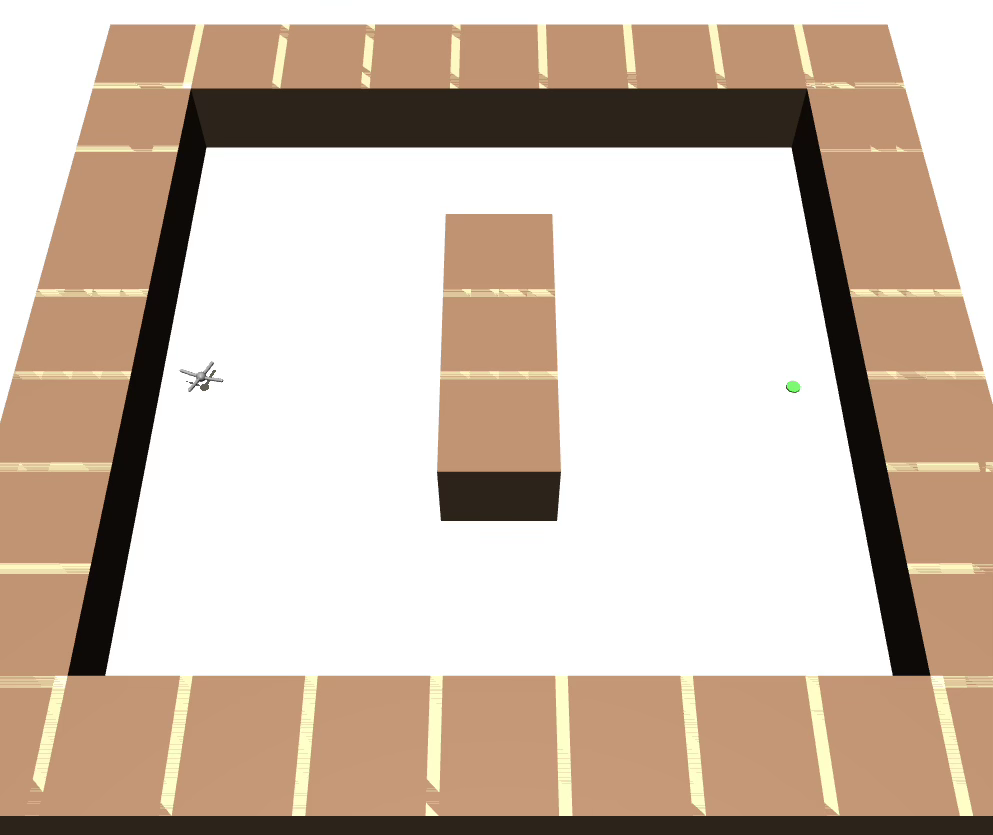} &
\includegraphics[height=0.353\linewidth]{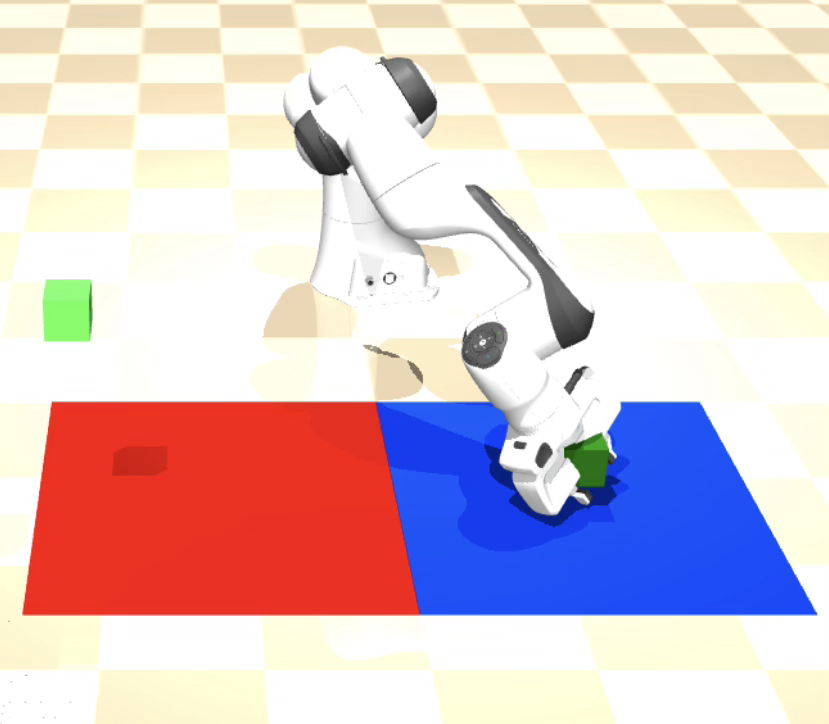} \\
\hline
\end{tabular}

\caption{
    Environments used for evaluation.
    \textit{Top:} Pure-exploration environments, with task rewards removed: Behaviour Suite DeepSea ($N=50$) \citep{osband_behaviour_2020}, Spiral PointMaze, and DMC Cheetah \citep{tassa2018deepmindcontrolsuite}.
    Together, these environments span discrete exploration, navigation, and high-dimensional continuous control.
    \textit{Bottom:} Complex-control environments: AntMaze and PickAndPlace \citep{zakka_mujoco_2025}.
    AntMaze requires navigating around obstacles to reach the goal, while PickAndPlace requires discovering coordinated manipulation behavior.
}
\label{fig:environments}
\end{figure}

\subsection{Pure Exploration}
Efficient reinforcement learning ultimately requires balancing exploration against exploitation, but this trade-off can obscure whether an exploration method is able to identify novelty in the first place.
We therefore first consider pure exploration, where task rewards are removed and behavior is driven entirely by the exploration signal.
Without an extrinsic objective specifying which states matter, a successful method should continually expand the region of the state space it visits, making state coverage a direct measure of its ability to identify and reach novel states.

We evaluate this ability across three environments of increasing complexity.
DeepSea uses a $50\times50$ grid with two actions and one-hot state encoding; the agent descends one row at each step.
In the spiral PointMaze, the agent starts at the center and chooses a two-dimensional displacement; a move that collides with a wall leaves the position unchanged.
DMC Cheetah provides a higher-dimensional continuous-control setting with 17-dimensional state observations and six-dimensional actions.
The exploration budgets are 500k steps for DeepSea and 5M steps for PointMaze and Cheetah.

\paragraph{Component ablations.}
The pure-exploration setting also allows us to isolate which parts of DEVOTE determine the quality of the exploration signal.
We first vary the RFN prior length scale over $\ell\in\{1,2.5,5,7.5\}$ to study the effect of prior complexity.
We then remove the RFN prior, soft resets, or TUD learning one at a time to isolate the contribution of each component.
The selected environment-specific length scale is otherwise shared across variants.

\subsection{High-Dimensional Control}
Both control tasks pair a shaped reward that teaches a basic skill with a sparse goal requiring exploration beyond the shaping-induced local optimum.
\paragraph{AntMaze.}
A Brax ant is placed in a MuJoCo maze with a $9\times9$ grid, with start and goal at opposite ends of the central row and a three-block wall obstructing the direct route.
Observations comprise 15 generalized positions, 14 generalized velocities, and the two-dimensional goal; actions have eight dimensions.
Episodes last at most 500 steps and terminate on success, but not on unhealthy states.
The reward is
\begin{equation}
    r=v_{\mathrm{goal}}-0.5(1-h)-0.1\|a\|_2^2
      -5\times10^{-4}\sum_i\operatorname{clip}(f_i,-1,1)^2+r_{\mathrm{succ}},
\end{equation}
where $v_{\mathrm{goal}}$ is torso velocity projected toward the goal, $h$ indicates a torso height in $[0.15,1.2]$ m, and $f_i$ are external contact forces.
A one-time reward of 2,500 is granted within one metre of the goal.
The signed velocity projection rewards movement toward the goal and penalizes movement away.
The direct reward gradient points toward the obstructing wall, making a detour necessary while retaining the learned locomotion skill.

\paragraph{PickAndPlace.}
A Franka Panda arm begins with a cube already grasped, isolating lifting and placement from grasp acquisition.
The cube starts at $(-0.25,0.65,0.07)$ m and the goal is at $(0.425,0.65,0.30)$ m.
The 27-dimensional observation contains arm joint positions and velocities ($7+7$), end-effector position ($3$), finger positions and velocities ($2+2$), and cube and goal positions ($3+3$); the eight-dimensional action controls the arm and gripper.
Episodes last at most 250 steps and terminate on success.
The reward is
\begin{equation}
    r=2\frac{\operatorname{clip}(z-0.03,0,0.12)}{0.12}-1
      -0.5(1-h_{\mathrm{grasp}})-0.1\|a\|_2^2+r_{\mathrm{succ}},
\end{equation}
where $h_{\mathrm{grasp}}$ indicates a gripper-to-cube distance below $0.12$ m.
A one-time reward of 1,250 is granted when the cube is within $0.05$ m of the goal.
The lifting term saturates at height $0.15$ m and provides no directional signal toward the displaced goal.
Both control tasks use 40M environment steps.

\paragraph{Coverage.}
Coverage is the fraction of eligible discretized bins visited at least once.
The coordinates are maze position for PointMaze and AntMaze, planar velocity for Cheetah, and cube position while the cube is grasped for PickAndPlace.
AntMaze divides each open maze cell into four bins and excludes walls.

\newpage
\section{Extended Results}

\subsection{Calibrating Epistemic Neural Networks}\label{app:regression}

\begin{figure}[htbp]
    \centering
    \includegraphics[width=\linewidth]{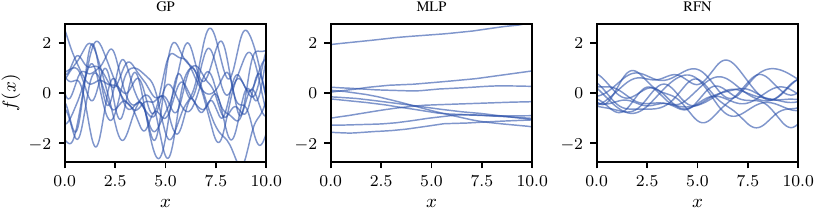}
    \caption{Samples from a GP prior, a randomly initialized ReLU MLP prior, and an RFN prior on the synthetic one-dimensional domain. The sampled MLP functions are approximately linear over this interval, whereas the RFN distributes variability throughout the domain. The GP and RFN use an RBF length scale of one.}
    \label{fig:prior-comparison}
\end{figure}

For uncertainty to guide exploration, unfamiliar inputs must remain distinguishable from those already supported by data.
The synthetic regression experiment shows how the choice of prior can undermine this distinction: both the bootstrapped ensemble and the ENN with a ReLU MLP prior become confident between the training-data clusters, despite inaccurate predictions there.
Figure~\ref{fig:prior-comparison} helps explain this behavior.
Over the illustrated domain, the sampled ReLU MLP priors are approximately linear, imposing strong correlations between distant inputs.
Fitting these simple functions near the observations can therefore suppress prior-induced variability even in the gaps between them.
The RFN prior instead distributes variability throughout the domain, with the length scale controlling how quickly function values decorrelate.
In the synthetic experiment, this construction preserves uncertainty between the observed clusters, where fitting the prior at the training points does not suffice to cancel its variation.

Table~\ref{tab:calibration_main} examines whether this improved distinction between observed and unfamiliar regions extends beyond the synthetic example.
Across the eight UCI datasets, the RFN ensemble with $\ell=0.25$ achieves a mean OOD-detection AUROC of $0.844$, compared with $0.749$ for the MLP-prior ensemble and $0.734$ for the bootstrapped ensemble (see~\ref{app:enn_cal_iml} for details).
This is the strongest mean OOD-detection result among the evaluated settings.
A shorter RFN length scale allows prior values to decorrelate closer to the observations, consistent with a sharper separation between inputs supported by the training data and those outside that distribution.

The error--uncertainty correlations in Table~\ref{tab:calibration_main} qualify this improvement.
The bootstrapped ensemble achieves the highest average correlation on the UCI datasets, whereas RFNs improve this correlation on the synthetic GP task.
Among the two RFN settings shown, $\ell=0.5$ yields stronger mean correlation, while $\ell=0.25$ yields better mean OOD detection.
The two metrics therefore reveal different strengths: distinguishing unfamiliar regions does not necessarily imply accurately ranking prediction errors.
For exploration, these results support RFNs as a way to preserve an informative uncertainty signal where observations are missing, without establishing uniformly better error calibration.

\begin{table*}[!htbp]
\vspace{1ex}
\renewcommand{\arraystretch}{1.2}
\small
\centering
\setlength{\tabcolsep}{3pt}
\begin{adjustbox}{max width=\linewidth}
\begin{tabular}{lllllllllll}
\toprule
Model & Concrete & Energy & Kin8nm & Naval & Protein & Power & Wine & Yacht & GP & Mean \\
\midrule
\rowcolor{lightgray}
\textbf{Boot-Ens} \\
\ AUROC OOD & 0.762 & \textbf{0.896} & 0.575 & 0.993 & 0.633 & 0.557 & 0.532 & 0.922 & 0.931 & 0.734 \\
\ $\rho(|f(x)-\mu(x)|,\sigma(x))$ & \textbf{0.365} & \textbf{0.487} & \textbf{0.104} & 0.617 & \textbf{0.158} & 0.005 & 0.134 & \textbf{0.565} & 0.410 & \textbf{0.304} \\
\midrule
\rowcolor{lightgray}
\textbf{ENN-Ens-MLP} \\
\ AUROC OOD & 0.869 & 0.816 & 0.537 & 1.000 & 0.632 & 0.608 & 0.621 & 0.907 & 0.995 & 0.749 \\
\ $\rho(|f(x)-\mu(x)|,\sigma(x))$ & \underline{0.308} & \underline{0.148} & 0.057 & 0.552 & 0.027 & -0.012 & \textbf{0.182} & 0.155 & 0.393 & 0.177 \\
\midrule
\rowcolor{lightgray}
\textbf{ENN-Ens-RFN ($\ell=0.25$)} \\
\ AUROC OOD & \textbf{0.924} & 0.847 & \textbf{0.603} & \textbf{1.000} & \textbf{0.779} & \textbf{0.907} & \textbf{0.720} & \textbf{0.976} & \textbf{0.997} & \textbf{0.844} \\
\ $\rho(|f(x)-\mu(x)|,\sigma(x))$ & 0.303 & 0.068 & 0.029 & \textbf{0.662} & \underline{0.039} & \textbf{0.015} & 0.140 & 0.147 & \underline{0.596} & 0.176 \\
\midrule
\rowcolor{lightgray}
\textbf{ENN-Ens-RFN ($\ell=0.5$)} \\
\ AUROC OOD & \underline{0.884} & \underline{0.856} & \underline{0.593} & \underline{1.000} & \underline{0.699} & \underline{0.802} & \underline{0.625} & \underline{0.956} & \underline{0.997} & \underline{0.802} \\
\ $\rho(|f(x)-\mu(x)|,\sigma(x))$ & 0.292 & 0.114 & \underline{0.079} & \underline{0.646} & 0.024 & \underline{0.012} & \underline{0.176} & \underline{0.184} & \textbf{0.681} & \underline{0.191} \\
\bottomrule
\end{tabular}
\end{adjustbox}
\vspace{1ex}
\caption{Calibration performance of ENN variants on regression benchmarks. We evaluate calibration across standard regression benchmarks and a synthetic GP task (RBF length scale $1$). The \textbf{best} performance is marked in bold and the \underline{second best} is underlined. The Mean column averages over the eight standard benchmarks (excluding GP).}
\label{tab:calibration_main}
\end{table*}

\subsection{Stabilizing Epistemic Value Functions}\label{app:ext_res_value_est}

An informative prior alone does not ensure reliable uncertainty estimates when learning value functions.
Under standard TD learning, the trainable network must fit the reward-dependent value while compensating for the fixed prior in both the current prediction and the bootstrapped target.
TUD separates reward fitting, prior cancellation, and residual bootstrapping into distinct objectives.
The policy evaluation experiments examine whether this separation allows uncertainty to decrease in frequently visited states while remaining informative elsewhere.

\begin{figure}[htbp]
    \centering
    \includegraphics[width=\linewidth]{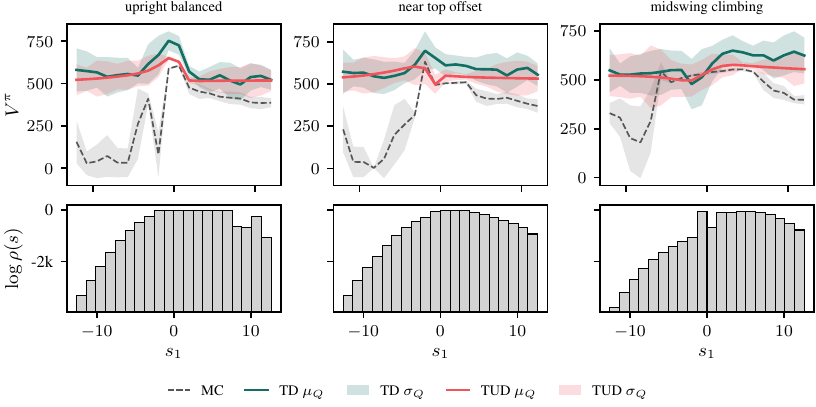}
    \caption{Fixed-policy cartpole evaluation with RFN length scale $\ell=1$. Columns show upright, near-top, and mid-swing anchors while pole angular velocity is varied and the remaining state coordinates and anchor action are held fixed. Top: Monte Carlo returns and TD/TUD predictions, with shaded bands showing one predicted standard deviation $\sigma_Q$. Bottom: estimated log visitation density. TUD reduces uncertainty in frequently visited regions while retaining uncertainty outside them; TD produces comparatively uniform uncertainty across the slices.}
    \label{fig:app-cartpole-slices}
\end{figure}

Figure~\ref{fig:app-cartpole-slices} makes this distinction visible by comparing the learned predictions with Monte Carlo returns and the policy's visitation density.
Along the three DMC cartpole state slices, TUD closely matches the Monte Carlo reference and reduces uncertainty in high-density regions.
Its uncertainty remains elevated in low-density regions, where observations provide less constraint on the prediction.
Standard TD learning instead produces comparatively uniform uncertainty across the slices, with little distinction between frequently visited and unfamiliar states, and slightly overestimates the reference values.
The difference is therefore not simply the overall magnitude of uncertainty, but whether it reflects where the policy has collected experience.

Both methods nevertheless predict the value poorly in parts of the low-density regions.
TUD improves the relationship between uncertainty and data availability without eliminating extrapolation error.
The shaded bands represent one predicted standard deviation, rather than calibrated confidence intervals, so they need not contain the Monte Carlo reference throughout each slice.

\begin{figure}[htbp]
    \centering
    \includegraphics[width=\linewidth]{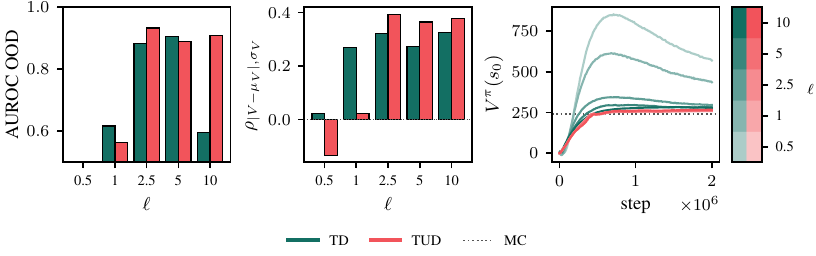}
    \caption{Fixed-policy cheetah evaluation across RFN prior length scales. Left: OOD-detection AUROC. Middle: error--uncertainty correlation. Right: predicted initial-state value over training compared with Monte Carlo returns. TUD improves both calibration metrics for $\ell\geq2.5$, while standard TD learning exhibits pronounced overestimation at small length scales. Results use three seeds.}
    \label{fig:app-cheetah-calibration}
\end{figure}

The DMC cheetah results in Figure~\ref{fig:app-cheetah-calibration} examine whether this behavior extends to a higher-dimensional task and how it depends on prior complexity.
For sufficiently large RFN length scales ($\ell\geq2.5$), TUD improves both OOD detection and the correlation between prediction error and uncertainty.
The initial-state predictions also expose pronounced overestimation under standard TD learning, particularly with small length scales.
These more rapidly varying priors make the interaction between prior compensation and reward fitting especially problematic in the joint update.

The improvements are not uniform across all length scales, showing that separating the learning objectives does not remove sensitivity to the prior.
Together with the cartpole slices, these results support the role of TUD in learning a useful epistemic value function: prior cancellation can reduce uncertainty where data are available, while separate reward fitting and residual bootstrapping improve value estimation without requiring uncertainty to vanish in unfamiliar states.

\begin{figure}[h]
    \centering
    \begin{subfigure}{\linewidth}
        \centering
        \includegraphics[width=\linewidth]{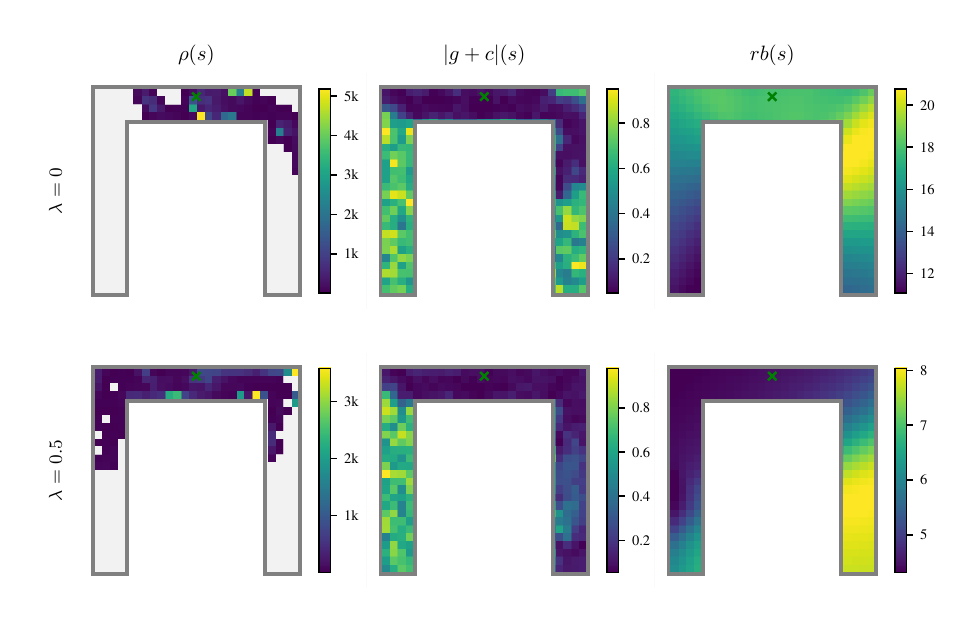}
        \caption{Early checkpoint - 250k steps. Both agents initially explore around the starting region.}
    \end{subfigure}
    \par\medskip
    \begin{subfigure}{\linewidth}
        \centering
        \includegraphics[width=\linewidth]{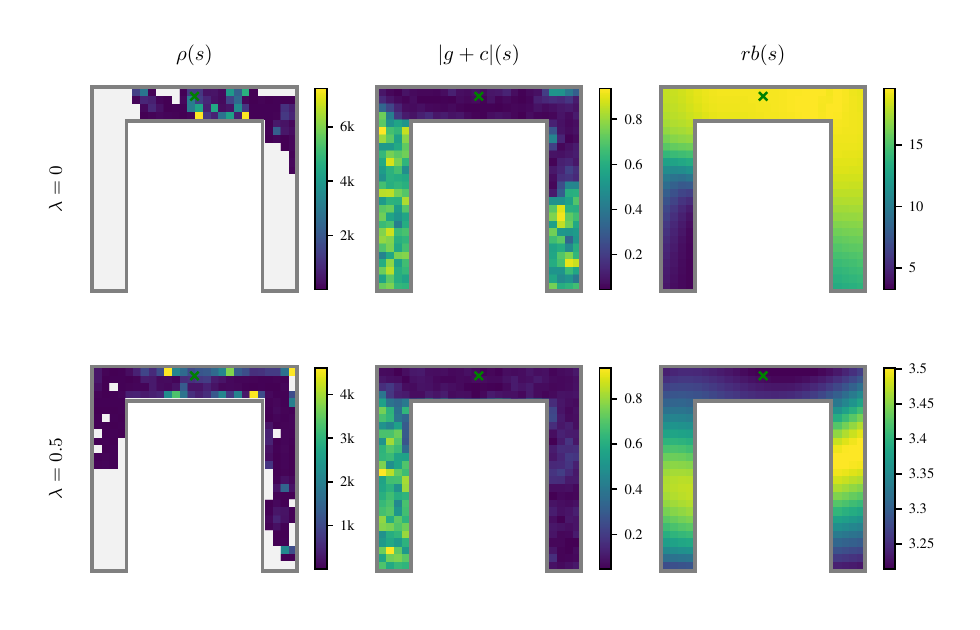}
        \caption{Later checkpoint - 500k steps. The agent with resets follows the novelty signal into the other corridor, while exploration without resets stalls.}
    \end{subfigure}
    \caption{U-shaped PointMaze diagnostics without resets ($\lambda=0$) and with resets ($\lambda=0.5$). Maps show visitation density $\rho(s)$, mean absolute residual $|r|(s)$, and mean residual bootstrap $rb(s)$. Each snapshot depicts a single run. Color scales are specific to each map, so the comparison concerns spatial structure rather than absolute color intensity across maps.}
    \label{fig:app-plasticity-snapshots}
\end{figure}

\subsection{Maintaining Plasticity in Optimistic Exploration}\label{app:plasticity_reg_res}

A useful novelty signal must continually change as the agent gathers experience.
As the corrector cancels the prior in visited regions, the remaining residual shifts the exploration objective toward unfamiliar states.
The residual bootstrap and exploration policy must keep adapting to this changing signal.
The U-shaped PointMaze experiment examines this requirement: after exploring the initial region, the agent must redirect its policy toward the other arm of the maze.

Figure~\ref{fig:app-plasticity-snapshots} shows how this process develops with and without soft resets.
The visitation density $\rho(s)$ records where the agent has collected experience, the residual magnitude $|r|(s)$ represents local novelty, and the residual bootstrap $rb(s)$ represents the novelty reachable through future interaction.
Reading these maps together reveals whether unfamiliar regions produce a signal that is propagated and followed by the agent.

At the earlier checkpoint, both agents explore around their initial state.
By the later checkpoint, their behavior has diverged.
With resets ($\lambda=0.5$), the residual bootstrap picks up the novelty signal in the other corridor, and visitation expands into that region.
Without resets ($\lambda=0$), the agent fails to track the shifting exploration objective and remains confined to the previously explored part of the maze.
The spatial diagnostics thus illustrate how exploration can stall even though unvisited regions remain: continued progress requires the networks to adapt as the location of novelty changes.

This behavior is consistent with the aggregate results in the main text, where stronger resets improve coverage and suppress the growth of dormant neurons.
The dormant-neuron count does not fully explain exploration performance, however: a moderate reset strength can reduce dormancy close to zero while the agent still underexplores.
The snapshots complement that diagnostic by showing how the learned signal and visitation evolve together.
They illustrate individual runs, while the coverage and dormant-neuron curves aggregate five seeds.
Together, these results support maintaining plasticity in both the critic and exploration actor so that an initially useful novelty signal can continue guiding behavior throughout training.

\subsection{Additional Exploration and Control Results}\label{app:main_res}

\begin{figure}[htbp]
    \centering
    \includegraphics[width=\linewidth]{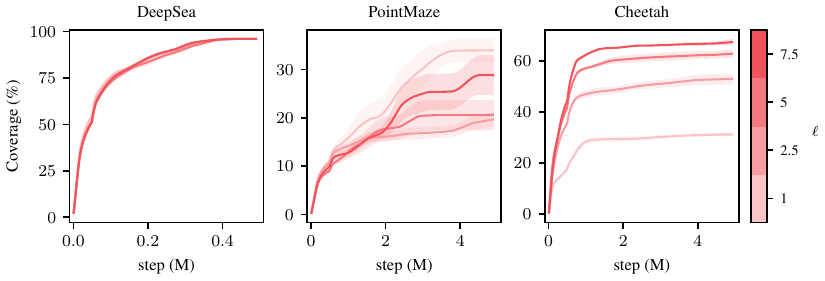}
    \caption{Pure-exploration performance of DEVOTE under varying RFN length scales $\ell\in\{1,2.5,5,7.5\}$, measured by state coverage. Prior complexity has little effect in the discrete DeepSea environment, while PointMaze favors the smaller length scale $\ell=1$ and Cheetah favors the larger length scale $\ell=7.5$.}
    \label{fig:length_scales_devote}
\end{figure}

\paragraph{Pure Exploration} The regression experiments show that prior complexity affects how uncertainty distinguishes familiar inputs from unfamiliar ones.
The length-scale sweep in Figure~\ref{fig:length_scales_devote} examines how this choice translates into exploration.
We vary the RFN length scale over $\ell\in\{1,2.5,5,7.5\}$, with smaller values producing more rapidly varying prior functions.
In DeepSea, the coverage curves are largely unchanged across the sweep.
This is consistent with the one-hot state encoding: the priors can distinguish individual states without relying on the spatial structure that matters in continuous domains.

In PointMaze and Cheetah, prior complexity instead strongly affects coverage, with the best-performing settings at $\ell=1$ and $\ell=7.5$, respectively.
We attribute this difference to the difficulty of fitting the prior across visited regions.
In the low-dimensional PointMaze, a more complex prior can retain variation outside the observed region while being canceled where data are available.
In the higher-dimensional Cheetah environment, fitting such a rapidly varying prior is harder, favoring a smoother prior that allows experience to suppress novelty in familiar states.
The sweep therefore suggests that an effective exploration prior must balance maintaining variation in unfamiliar regions with being learnable in visited ones; increasing prior complexity alone does not consistently improve exploration.

\paragraph{Complex Control} The control tasks examine whether exploration remains effective when the agent must also exploit an extrinsic reward.
In AntMaze, the shaped reward teaches goal-directed locomotion, but the agent must discover a detour around a wall to reach the goal.
In PickAndPlace, the dense lift reward encourages raising the cube without directing it toward the displaced, elevated target.
Both tasks therefore require the agent to explore beyond the behavior already encouraged by reward shaping.

\begin{figure}[htbp]
    \centering
    \includegraphics[width=\linewidth]{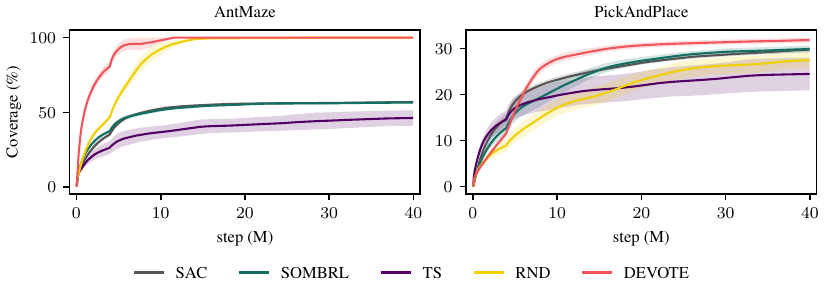}
    \caption{State coverage in AntMaze and PickAndPlace, complementing the normalized returns in the main text. Coverage measures the fraction of visited bins in ant $x$--$y$ position and grasped-cube $x$--$y$--$z$ position, respectively. DEVOTE expands coverage fastest in both environments. RND also reaches full coverage in AntMaze, but achieves lower coverage than SAC in PickAndPlace.}
    \label{fig:app-control-coverage}
\end{figure}

Figure~\ref{fig:app-control-coverage} complements the main-text return curves by showing how far this exploration extends.
DEVOTE expands coverage fastest in both environments, consistent with its stronger task performance.
In AntMaze, RND also reaches full coverage and is the only baseline reported to solve the sparse-reward task.
In PickAndPlace, however, RND covers less of the cube-position space than SAC.
Its success in AntMaze therefore does not transfer consistently to the manipulation setting.

Read alongside the return curves, these results distinguish acquiring the initially rewarded behavior from exploring beyond it.
Higher coverage indicates that the agent reaches a broader set of positions, while return measures whether its behavior advances the task.
DEVOTE's gains in both quantities support its ability to combine novelty-driven exploration with the skills learned from extrinsic reward.

\end{document}